\documentclass[twocolumn,10pt]{article}
\usepackage[
top=1.78cm,
bottom=1.78cm,
left=1.4cm,
right=1.4cm,
headsep=0cm,
]{geometry}
\usepackage{etoolbox}
\usepackage[T1]{fontenc}
\usepackage{cite}
\usepackage{amsmath,amssymb,amsfonts}
\usepackage{algorithmic}
\usepackage{graphicx}
\usepackage{placeins}
\usepackage{setspace}
\usepackage{enumitem}
\usepackage{titlesec}
\usepackage{textcomp}
\usepackage[dvipsnames]{xcolor}
\usepackage[binary-units=true]{siunitx}
\usepackage[export]{adjustbox}
\usepackage{blindtext}
\usepackage{booktabs}
\usepackage{tikz}
\usepackage{multirow}
\usepackage{mwe,tikz}\usepackage[percent]{overpic}
\usepackage[font=footnotesize]{subcaption}
\usepackage[font=small, skip=5pt]{caption}
\usepackage[affil-it]{authblk}
\usetikzlibrary{shapes}
\usetikzlibrary{shapes.misc}
\tikzset{cross/.style={cross out, draw=black, minimum size=2*(#1-\pgflinewidth), inner sep=0pt, outer sep=0pt},
cross/.default={0.6mm}}

\DeclareSymbolFont{matha}{OML}{txmi}{m}{it}
\DeclareMathSymbol{\varv}{\mathord}{matha}{118}
\def\BibTeX{{\rm B\kern-.05em{\sc i\kern-.025em b}\kern-.08em
    T\kern-.1667em\lower.7ex\hbox{E}\kern-.125emX}}

\renewrobustcmd{\bfseries}{\fontseries{b}\selectfont}
\renewrobustcmd{\boldmath}{}
\newrobustcmd{\B}{\bfseries}

\AtBeginEnvironment{tabular}{\footnotesize}
\AtBeginEnvironment{abstract}{\itshape}
\AtBeginEnvironment{thebibliography}{\fontsize{8.5pt}{9.75pt}\selectfont}
\apptocmd{\thebibliography}{\setlength{\itemsep}{-4pt}}{}{}

\setlist{nosep}

\renewcommand{\thesection}{\Roman{section}}
\renewcommand{\thesubsection}{\Alph{subsection}}
\renewcommand{\thesubsubsection}{\arabic{subsubsection}}
\titleformat{\section}{\normalfont\large\scshape\centering}{\thesection.}{0.5em}{}
\titleformat{\subsection}{\normalfont\itshape}{\thesubsection.}{0.5em}{}
\titleformat{\subsubsection}[runin]{\normalfont\itshape}{\thesubsubsection)}{0.5em}{}[:]
\titlespacing{\section}{0pt}{1.5ex plus 0.5ex minus .2ex}{1.0ex plus .2ex}
\titlespacing{\subsection}{0pt}{1.1ex plus 0.5ex minus .2ex}{0.5ex plus .2ex}
\titlespacing{\subsubsection}{\parindent}{0pt}{0.5em}
\makeatletter
\renewcommand{\p@section}{}
\renewcommand{\p@subsection}{\thesection-}
\renewcommand{\p@subsubsection}{\thesection-\thesubsection.}
\makeatother

\definecolor{light-gray}{gray}{0.95}
\AtBeginEnvironment{snugshade*}{\vspace{-\FrameSep}}
\AfterEndEnvironment{snugshade*}{\vspace{-\FrameSep}}

\makeatletter
\def\footnoterule{\relax%
  \kern-5pt
  \hbox to \columnwidth{\vrule width 0.45\columnwidth height 0.4pt\hfill}
  \kern4.6pt
  }
\makeatother

\makeatletter
\newcommand\semiHuge{\@setfontsize\semiHuge{22.72}{29.5}}
\makeatother
\makeatletter
\patchcmd{\@maketitle}{\@title}{\vspace{-1cm}\semiHuge \@title}{}{}
\patchcmd{\@maketitle}{\@author}{\normalsize\@author}{}{}
\makeatother
\begin{document}
\title{Computer Vision Tool for Detection, Mapping and Fault Classification of PV Modules in Aerial IR Videos}

\author[1]{Lukas Bommes}
\author[1]{Tobias Pickel}
\author[1]{Claudia Buerhop-Lutz}
\author[1]{Jens Hauch}
\author[1,2]{Christoph Brabec}
\author[1]{Ian Marius Peters}
\affil[1]{Forschungszentrum Jülich GmbH, Helmholtz-Institute Erlangen-Nuremberg for Renewable Energies (HI ERN)}
\affil[2]{Institute Materials for Electronics and Energy Technology, Universität Erlangen-Nürnberg (FAU)\par\normalsize\normalfont{Correspondence to i.peters@fz-juelich.de}}

\date{}

\maketitle
\thispagestyle{empty} 

\begin{abstract}
\begin{spacing}{0.9}
Increasing deployment of photovoltaics (PV) plants demands for cheap and fast inspection. A viable tool for this task is thermographic imaging by unmanned aerial vehicles (UAV). In this work, we develop a computer vision tool for the semi-automatic extraction of PV modules from thermographic UAV videos. We use it to curate a dataset containing $4.3$ million IR images of $107842$ PV modules from thermographic videos of seven different PV plants. To demonstrate its use for automated PV plant inspection, we train a ResNet\nobreakdash-$50$ to classify ten common module anomalies with more than \SI{90}{\percent} test accuracy. Experiments show that our tool generalizes well to different PV plants. It successfully extracts PV modules from $512$ out of $561$ plant rows. Failures are mostly due to an inappropriate UAV trajectory and erroneous module segmentation. Including all manual steps our tool enables inspection of \SI{3.5}{\mega\watt}$_\textrm{p}$ to \SI{9}{\mega\watt}$_\textrm{p}$ of PV installations per day, potentially scaling to multi-gigawatt plants due to its parallel nature. While we present an effective method for automated PV plant inspection, we are also confident that our approach helps to meet the growing demand for large thermographic datasets for machine learning tasks, such as power prediction or unsupervised defect identification.
\end{spacing}
\end{abstract}



\section{Introduction}

Deployment of solar photovoltaics (PV) has increased exponentially in the past years. At the end of $2019$, globally installed capacity reached \SI{586}{\giga\watt}$_\textrm{p}$ \cite{bpstatistic.2020}. Many PV plants contain defective PV modules which pose safety hazards and reduce power output, yield and as a consequence, the profitability of the plant. Defects occur during manufacturing, installation or due to aging. To identify defective modules PV plants need to be inspected regularly.

A valuable tool for defect identification in PV modules is thermographic imaging which uses a thermal IR camera to visualize defects based on their increased temperature. To speed up the inspection process thermography is typically performed by unmanned aerial vehicles (UAV) \cite{Buerhop.2012, Buerhop.2012b, Scheuerpflug.2014, Quater.2014}. Many works have explored the use of UAVs for PV plant inspection. A high-level overview of the inspection process and the challenges involved is given in \cite{Niccolai.2019b, Kumar.2018}. \cite{Gallardo-Saavedra.2018} compares available camera and drone technologies and \cite{Bizzarri.2019} performs an economical analysis. \cite{Gallardo-Saavedra.2018b, Leva.2015} analyze the influence of the image resolution on the detectability of defects.

UAV thermography of PV plants with millions of modules produces so many images that manual sighting is infeasible. This raises the need for image processing tools which automatically detect PV modules in each image and identify thermal anomalies. To enable repairs or exchange of defective modules the automated processing tool needs to further determine the exact location of each module in the plant. Instead of taking individual images at predetermined positions, we simply fly along each row of the PV plant and acquire videos. This renders expensive and time consuming flight planning unnecessary and allows for faster inspection on-site. However, it increases the amount of data as each PV module occurs in multiple consecutive video frames. It further introduces perspective distortion and other artefacts, such as sun reflections, which need to be handled by the processing tool to make the images usable for downstream anomaly classification and other machine learning algorithms. The large number of acquired thermographic images is key to accurate anomaly classification as some anomalies are very seldom and machine learning algorithms used for anomaly classification require many examples to achieve high accuracy and good generalization.

\begin{figure}[tbp]
  \centering
  \begin{overpic}[scale=1.0]{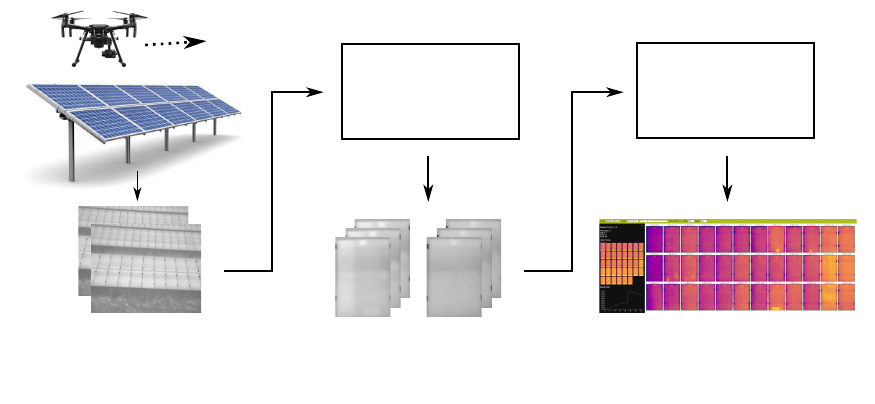}
  \put (39,35) {\parbox{18.18mm}{\footnotesize \centering PV Module\\Extraction}}
  \put (72.5,35) {\parbox{18.18mm}{\footnotesize \centering Anomaly\\Classification}}
  \put (2,4) {\parbox{25mm}{\footnotesize \centering Thermal IR videos\\of the PV plant}}
  \put (34,4) {\parbox{25mm}{\footnotesize \centering IR patches\\of PV modules}}
  \put (79,5) {\footnotesize Report}
  \end{overpic}
  \caption{High-level overview of our tool for semi-automatic inspection of PV plants using thermographic videos acquired by an UAV.}
  \label{fig:high_level_overview}
\end{figure}
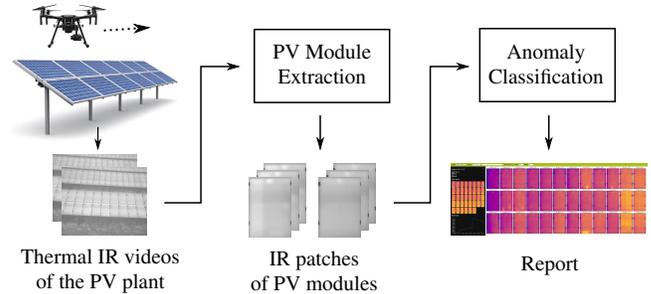

In this work we develop such an image processing tool for the semi-automatic extraction and localization of PV modules in UAV thermographic videos of large-scale PV plants (see fig. \ref{fig:high_level_overview}). It can be used to automate inspection of PV plants and to curate large datasets for downstream machine-learning tasks. While there are several works on automated PV plant inspection systems \cite{Carletti.2019, Grimaccia.2017, Grimaccia.2018, Niccolai.2019, Arenella.2017, Henry.2020b, Addabbo.2018, Deitsch.2016}, they rely heavily on classic image processing techniques, such as intensity thresholding (see tab. \ref{tab:related_works}). These techniques are based on heuristics, need extensive manual tuning, do not generalize well and are not very accurate. Further, many of the related works can distinguish at most three different thermal anomalies or perform only a binary classification. First works have shown promising results using deep learning for these tasks \cite{Dunderdale.2020, Diaz.2020}. Following this recent trend, we use the Mask R\nobreakdash-CNN instance segmentation framework \cite{He.2017} to robustly extract PV modules from thermal IR videos. A ResNet\nobreakdash-$50$ deep convolutional classifier \cite{He.2016} is used for fine-grained classification of ten thermal anomalies. Further, we exploit the large redundancy and temporal context present in the video data to efficiently build a large-scale dataset of thermographic images of PV modules for downstream machine learning tasks. To summarize, our contributions are as follows:
\begin{itemize}
\item A tool for semi-automatic extraction and localization of PV modules in UAV thermographic videos of large-scale PV plants which can be used for automated plant inspection and to curate large datasets for downstream machine-learning tasks.
\item A dataset of $4.3$ million thermographic images of $107842$ PV modules from seven PV plants with fine-grained labels of ten common thermal anomalies.
\item Training and evaluation of a ResNet\nobreakdash-$50$ classifier on our dataset.
\item A quantitative analysis of generalization ability, processing time and failure cases of our tool.
\end{itemize}



\section{Related Works}
\label{sec:related_works}

The following is an overview of related methods for semi\nobreakdash-automatic thermographic PV plant inspection by UAVs. We compare them in terms of module detection, thermal anomaly detection and localization of modules in the plant. Tab. \ref{tab:related_works} summarizes methods and dataset sizes of the related works.

\subsection{PV Module Detection}

Most works employ classic computer vision algorithms to detect PV modules in both visual and thermographic images. The most popular method used by \cite{Grimaccia.2017, Grimaccia.2018, Niccolai.2019, Kim.2016, Kim.2017, Deitsch.2016} is binary thresholding of image intensities to obtain segmentation masks of the PV modules. \cite{Diaz.2020} detects rectangular candidate contours by thresholding, extracts texture features and classifies them with a Support Vector Machine (SVM). Other works find edges of PV modules using morphological operations \cite{Aghaei.2016, Wu.2017} or the Hough transform \cite{Arenella.2017, Carletti.2019}. More exotic techniques are template matching \cite{Addabbo.2018} and maximally stable extremal regions \cite{Henry.2020b}. Main issue of all these works is their reliance on classic image processing which is based on manual priors and heuristics, needs extensive manual tweaking of hyper parameters and generalizes poorly to unseen imagery.

Deep learning overcomes these problems and is applied to PV module detection by \cite{Zhang.2019, Greco.2020, Diaz.2020}. \cite{Zhang.2019} performs semantic segmentation with a combination of a ResNet\nobreakdash-34 \cite{He.2016} and a U\nobreakdash-Net \cite{Ronneberger.2015}. A weakness of semantic segmentation is that it does not distinguish between individual PV modules. \cite{Greco.2020} employs the YOLO object detector \cite{Redmon.2016} which does not have this problem. However, it suffers from the imprecise representation of PV modules by bounding boxes instead of segmentation masks. Similar to our work \cite{Diaz.2020} solves both problems by utilizing the Mask R\nobreakdash-CNN instance segmentation model. It outputs an individual segmentation masks for each PV module which allow for accurate localization of PV modules in thermographic images.

\subsection{Thermal Anomaly Detection}

Similar to the PV module detection many works \cite{Aghaei.2015, Arenella.2017, Henry.2020b, Grimaccia.2017, Grimaccia.2018} use binary thresholding to segment hot regions of PV modules in thermographic images which correspond to thermal anomalies. The works in \cite{Alsafasfeh.2018, Carletti.2019} iteratively grow segmentation masks of hot spots starting from local intensity maxima. In \cite{Addabbo.2018} hot spots are found by template matching. Another approach is to extract features, such as mean and standard deviation, for each PV module and finding outliers with statistical tests \cite{Deitsch.2016} or by comparing with neighbouring modules \cite{Kim.2017}.


Several recent works explore deep learning for anomaly detection to overcome the limitations of classic image processing \cite{Oliveira.2019, Pierdicca.2018, Dunderdale.2020}. In \cite{Oliveira.2019} a segmentation model based on VGG\nobreakdash-$16$ is used to segment three different anomalies directly in the thermographic image. VGG\nobreakdash-$16$ is also used by \cite{Pierdicca.2018} to classify whether an image contains an anomalous module or not. Problem of this method is the inability to accurately localize the anomalous module. In \cite{Dunderdale.2020} four different anomalies are classified using MobileNet and VGG\nobreakdash-$16$. The authors find that both deep learning methods outperform a SVM and a Random Forest classifier using SIFT features.

Problem of the current methods is that the list of anomalies classified is by no means complete. Further, small datasets with only $360$ to $3336$ images are used.

Similar to \cite{Dunderdale.2020} we utilize a deep convolutional classifier, in our case ResNet\nobreakdash-$50$. However, we obtain a significantly larger anomaly classification dataset with more than $450000$ images and perform a much more fine-grained classification of ten thermal anomalies. In addition, we employ majority voting over subsequent video frames to enhance classification accuracy.

\subsection{Localization of PV Modules in the Plant}

To localize PV modules in the PV plant \cite{Aghaei.2016, Grimaccia.2017, Grimaccia.2018} create panorama images of each row, detect modules and assign an ID to each module. This way, module locations are defined relative to other modules. \cite{Niccolai.2019} uses the same technique and additionally matches each row panorama to a CAD plan by means of GPS positions. Problematic is the need for an accurate flight path with specified overlap of individual images which makes the UAV operation more complicated. Further, CAD files are not always available and the format can vary for different PV plants.

Several works \cite{Lee.2019, Zefri.2018, Tsanakas.2017} create an orthophoto of the entire PV plant from a higher altitude. This requires nadiral images with a suitable overlap which may not always be feasible in case of nearby power lines, streets or train tracks. Spatial resolution of a high-altitude image is low making fine-grained anomaly classification of PV modules difficult.

Other works \cite{Addabbo.2018, Nisi.2016} use direct georeferencing to estimate the GPS position of each PV module in the image. This requires an expensive Real Time Kinematics system to accurately estimate the UAVs position.

In \cite{Henry.2020} GPS positions of the video frames containing an anomalous PV module are marked on a map. While this is straightforward it still requires manual localization of the anomalous module within the frame.

Our work uses relative mapping similar to \cite{Aghaei.2016, Grimaccia.2017, Grimaccia.2018}. Instead of creating a panorama, we encode the spatial relationship of PV modules in a graph that is matched with a standardized \emph{plant file} containing module identifiers. This allows for easy integration of other data modalities, such as electrical measurements. The plant file needs to be created only once for each plant which saves time when inspecting the same plant multiple times. We further do not require nadiral images or a specific overlap of adjacent frames and a standard GPS receiver is sufficient. This reduces cost and allows for a more flexible operation of the UAV.

\begin{table*}[htb]
\centering
\caption{Comparison of related works on PV module detection and thermal anomaly detection in aerial IR images of PV plants. F$1$-scores are taken from the original works and are not directly comparable due to different test datasets and different definitions of the F$1$-score (pixel-based, bounding box-based, choice of IoU threshold). A unification is out of the scope of this work. F$1$-scores defined in the same way as in our work are demarked with a \textsuperscript{$\dagger$}.}
\label{tab:related_works}
\begin{tabular}{
>{\raggedright\arraybackslash}p{1.2cm}
>{\raggedright\arraybackslash}p{2.3cm}
>{\raggedright\arraybackslash}p{3.8cm}
>{\raggedright\arraybackslash}p{1.0cm}
l
>{\raggedright\arraybackslash}p{4.0cm}
l
>{\raggedright\arraybackslash}p{2.5cm}
l
}
\toprule
Work & Test (train) dataset & \multicolumn{3}{c}{Module detection} & \multicolumn{4}{c}{Anomaly detection}\\\cmidrule(lr){3-5}\cmidrule(lr){6-9}
 & Images / Modules / Plants & Method & Type & F1/\% & Method & \multicolumn{2}{l}{Anomaly classes} & F1/\%\\\midrule
\cite{Henry.2020b} & 20 / 240 / 1 & Region proposal by Maximally Stable Extremal Regions (MSER) + filtering by size & Boxes & n.a. & Segmentation by binary thresholding & 1 & Hot spot & n.a.\\

\cite{Arenella.2017} & 1171 / -- / 1 & Edge extraction by Hough transform + postprocessing & Lines & n.a. & Segmentation by binary thresholding & 1 & Hot spot & 59.0\\

\cite{Grimaccia.2017, Grimaccia.2018, Niccolai.2019} & 34 / -- / 1 & Segmentation by binary thresholding in HSV-space & Mask & n.a. & Segmentation by binary thresholding with two thresholds + classification heuristics & 3 & Hot spot, hot substring, hot module & 98.8\textsuperscript{$\dagger$}\\

\cite{Deitsch.2016} & 37 / 1544 / 2 & Segmentation by binary thresholding with adaptive threshold & Mask & 92.8 & Feature extraction + classification with Grubb's test and Dixon's Q test & 3 & Hot spot, hot substring, hot module & 93.9\textsuperscript{$\dagger$}\\ 

\cite{Kim.2016, Kim.2017} & 3 / 204 / 1 & Segmentation by binary thresholding + morphological operations & Mask & 95.8 & Feature extraction (mean \& std) + comparison with neighbouring modules & 3 & Hot spot, hot substring, hot module & 92.9\textsuperscript{$\dagger$}\\

\cite{Addabbo.2018} & 270 / -- / 1 & Template matching & Boxes & 83.0 & Template matching & 1 & Hot spot & 75.0\textsuperscript{$\dagger$}\\

\cite{Carletti.2019} & -- / 14215 / >1 & Canny edge detection + Hough transform & Lines & 87.0 & Segmentation by water filling algorithm + temporal tracking with majority voting & 1 & Hot spot & 72.0\\


\cite{Diaz.2020} & test: 20 / -- / 3 \newline train: 80 / -- / 3 & Rectangle extraction by adaptive thresholding + SVM classifier on texture features & Boxes + Masks & 98.3 & -- & -- & -- & --\\

\cite{Dunderdale.2020} & test: 77 / -- / 3 \newline train: 306 / -- / 3 & -- & -- & -- & SIFT feature extraction + Random Forest classifier & 4 & Sh, Sp, Mp, Cs+ (see fig. \ref{fig:fault_classification_dataset_examples}) & 77.2\textsuperscript{$\dagger$}\\\midrule

\cite{Diaz.2020} & test: 20 / -- / 3 \newline train: 80 / -- / 3 & DL instance segmentation (Mask R-CNN) + postprocessing & Boxes + Masks & 98.9 & -- & -- & -- & --\\

\cite{Greco.2020} & test: -- / 14499 / >1 \newline train: -- / 36000 / >1 & DL object detection (YOLOv3) & Boxes & 95.0 & -- & -- & -- & --\\

\cite{Zhang.2019} & test: 19 / -- / 1 \newline train: 216 / -- / 1 & DL semantic segmentation (ResNet-34 + U-Net) & Mask & 97.1 & -- & -- & -- & --\\

\cite{Oliveira.2019} &  -- / -- / 1 & -- & -- & -- & Segmentation by VGG-16 based DL model & 3 & Hot spot, hot substring, hot string & n.a.\\

\cite{Pierdicca.2018} & test: 318 / -- / 1 \newline train: 1304 / -- / 1 & -- & -- & -- & DL classification (VGG-16) of entire video frame & 1 & Binary & 75.0\\

\cite{Dunderdale.2020} & test: 77 / -- / 3 \newline train: 306 / -- / 3 & -- & -- & -- & DL classification (MobileNet, VGG-16) & 4 & Sh, Sp, Mp, Cs+ (see fig. \ref{fig:fault_classification_dataset_examples}) & 89.5\textsuperscript{$\dagger$}\\\midrule

\end{tabular}
\end{table*}


\section{Video Dataset}
\label{sec:video_dataset}

\begin{figure}[htbp]
     \centering
     \begin{subfigure}[b]{0.24\columnwidth}
         \centering
         \begin{overpic}[width=\textwidth]{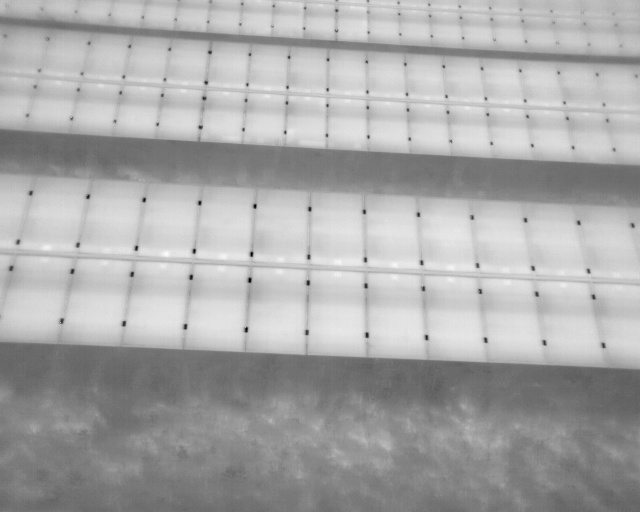}
          \put (5,5) {Plant A}
          \end{overpic}
     \end{subfigure}
     \begin{subfigure}[b]{0.24\columnwidth}
         \centering
         \begin{overpic}[width=\textwidth]{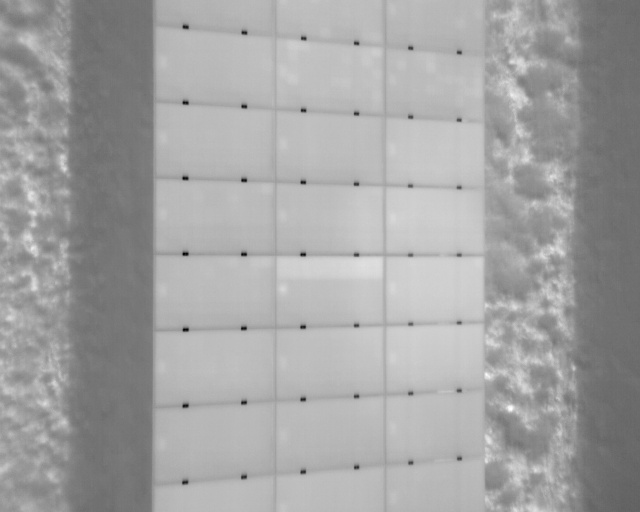}
          \put (5,5) {B}
          \end{overpic}
     \end{subfigure}
     \begin{subfigure}[b]{0.24\columnwidth}
         \centering
         \begin{overpic}[width=\textwidth]{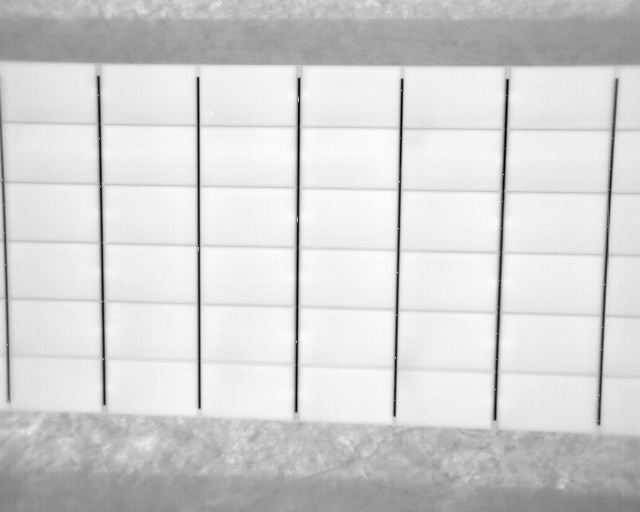}
          \put (5,5) {C}
          \end{overpic}
     \end{subfigure}
     \begin{subfigure}[b]{0.24\columnwidth}
         \centering
         \begin{overpic}[width=\textwidth]{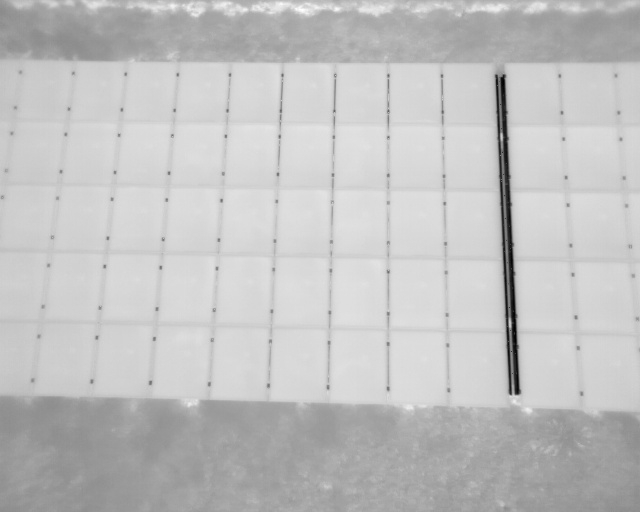}
          \put (5,5) {D}
          \end{overpic}
     \end{subfigure}
     \par\medskip
     \begin{subfigure}[b]{0.24\columnwidth}
         \centering
         \begin{overpic}[width=\textwidth]{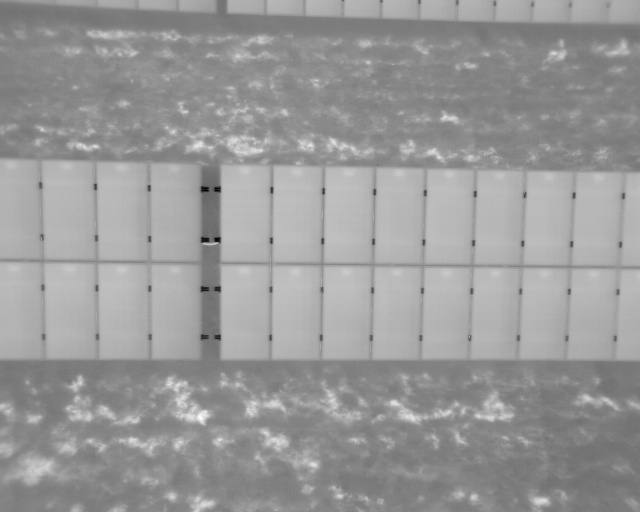}
          \put (5,5) {E}
          \end{overpic}
     \end{subfigure}
     \begin{subfigure}[b]{0.24\columnwidth}
         \centering
         \begin{overpic}[width=\textwidth]{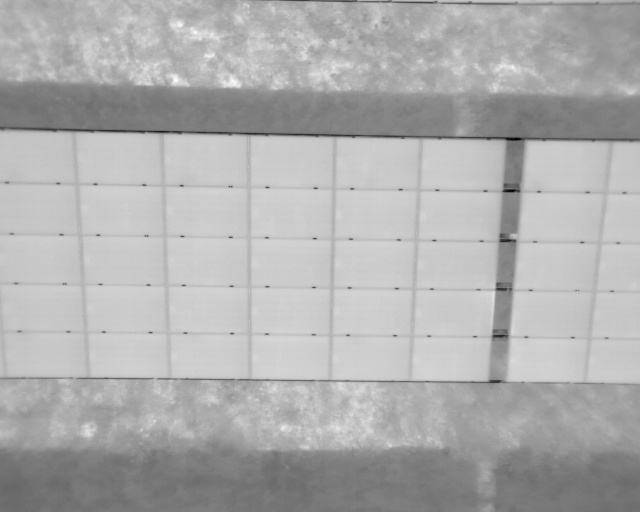}
          \put (5,5) {F}
          \end{overpic}
     \end{subfigure}
     \begin{subfigure}[b]{0.24\columnwidth}
         \centering
         \begin{overpic}[width=\textwidth]{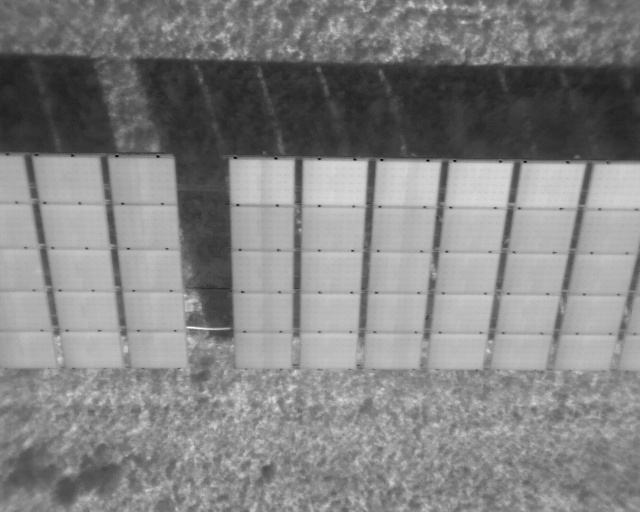}
          \put (5,5) {G}
          \end{overpic}
     \end{subfigure}
        \caption{Example video frames of the seven PV plants in our dataset.}
        \label{fig:plant_pictures}
\end{figure}

For this work we acquire thermographic videos of seven utility-scale PV plants containing a combined $122865$ PV modules (ranging from $2850$ to $35360$ modules per plant). As can be seen in fig. \ref{fig:plant_pictures} the plants in our dataset cover a variety of row layouts, module sizes, module orientations and module technologies. Plant D comprises of thin-film modules while the others use crystalline silicon modules. In total our dataset contains $8$ hours of video footage ($231172$ frames) with on average $21.8$ PV modules per frame. Videos were acquired by a UAV of type DJI Matrice $210$ and a DJI Zenmuse XT$2$ camera which has a resolution of $640 \times 512$ pixels and a frame rate of \SI{8}{\hertz}. Acquisition took place under clearsky conditions and solar irradiance above \SI{700}{\watt\per\square\meter}.

\section{PV Module Extraction}
\label{sec:pv_module_extraction}

\begin{figure*}[htbp]
  \definecolor{custom-green}{rgb}{0,0.803921569,0}
  \centering
  \begin{overpic}[scale=1.0, abs, unit=1mm, tics=5]{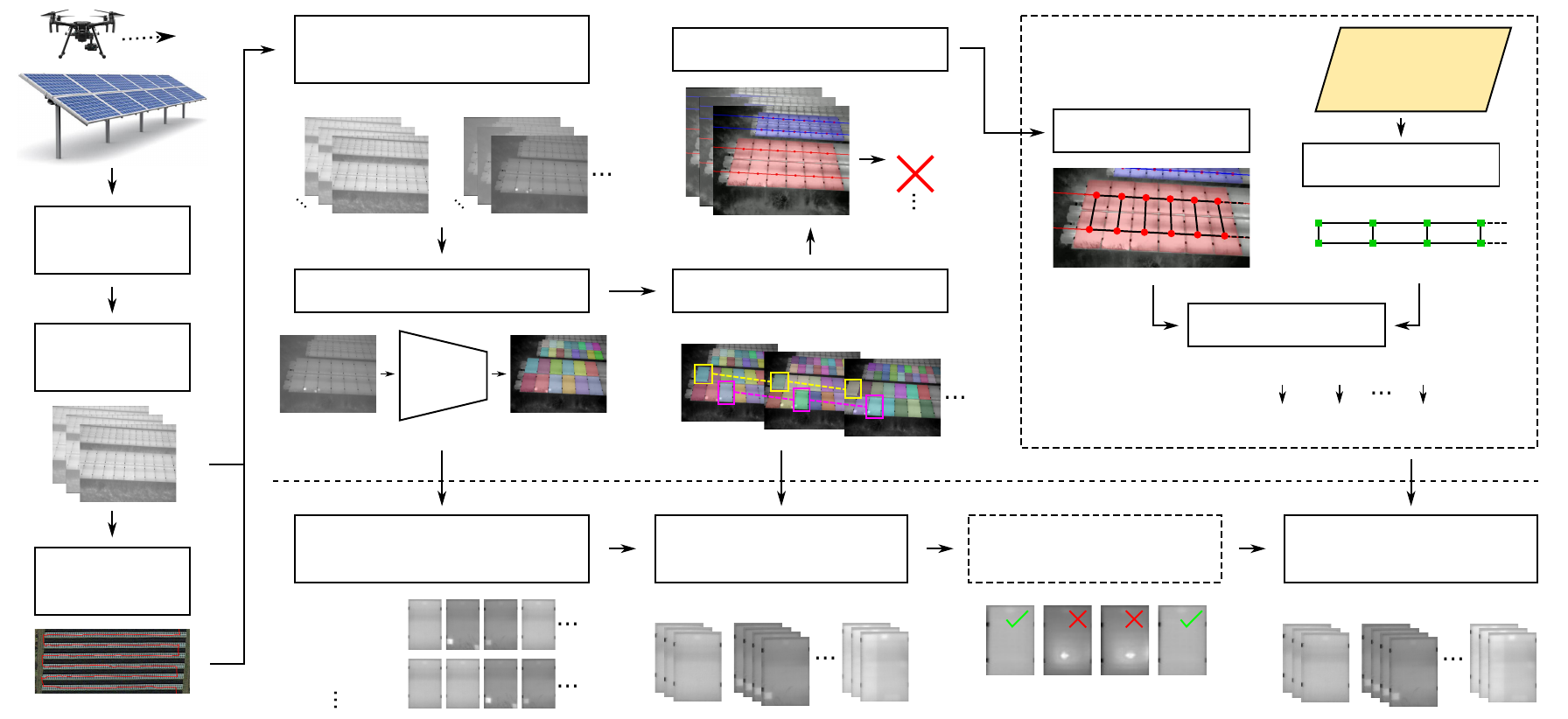}
  \put (3.930, 59.514) {\parbox[t][8mm][c]{18.205mm}{\footnotesize \centering Thermal IR\\video file}}
  \put (3.930, 45.876) {\parbox[t][8mm][c]{18.205mm}{\footnotesize \centering Split video\\in frames}}
  \put (3.930, 19.763) {\parbox[t][8mm][c]{18.205mm}{\footnotesize \centering Extract GPS\\coordinates}}
  \put (34.124, 81.334) {\parbox[t][8mm][c]{34.363mm}{\footnotesize \centering Group frames\\by PV plant row}}
  \put (32, 61.25) {\footnotesize $t_0$}
  \put (36, 57) {\footnotesize $t_1$}
  \put (50.5, 61.25) {\footnotesize $t_3$}
  \put (54.5, 57) {\footnotesize $t_4$}
  \put (37, 70.25) {\footnotesize Row $1$}
  \put (55.25, 70.25) {\footnotesize Row $2$}
  \put (34.124, 51.9) {\parbox[t][5.2mm][c]{34.363mm}{\footnotesize \centering Segment PV modules}}
  \put (46.310, 45.186) {\parbox[t][10.642mm][c]{10.332mm}{\footnotesize \centering Mask\\R-CNN}}
  \put (34.6, 32.7) {\footnotesize Frame}
  \put (61.3, 32.7) {\footnotesize Masks}
  \put (34.124, 23.477) {\parbox[t][8mm][c]{34.363mm}{\footnotesize \centering Extract rectified image\\patches of modules}}
  \put (35, 11) {\footnotesize Frame $t_0$}
  \put (35, 4.5) {\footnotesize Frame $t_1$}
  \put (78.061, 80.2) {\parbox[t][5.2mm][c]{32.107mm}{\footnotesize \centering Filter front row}}
  \put (104, 69) {\footnotesize 2bc}
  \put (104, 65.75) {\footnotesize fa3}
  \put (104, 62.5) {\footnotesize 5c7}
  \put (78.061, 52.282) {\parbox[t][5.2mm][c]{32.107mm}{\footnotesize \centering Track PV modules}}
  \put (83.5, 44.4) {\footnotesize $t_0$}
  \put (93, 43.45) {\footnotesize $t_1$}
  \put (102, 42.5) {\footnotesize $t_2$}
  \put (81, 41.25) {\footnotesize \color{white} 2bc}
  \put (89, 40.5) {\footnotesize \color{white} 2bc}
  \put (98, 39.5) {\footnotesize \color{white} 2bc}
  \put (80, 35) {\footnotesize \color{white} fa3}
  \put (89, 34.1) {\footnotesize \color{white} fa3}
  \put (101, 33.2) {\footnotesize \color{white} fa3}
  \put (75.968, 23.477) {\parbox[t][8mm][c]{29.533mm}{\footnotesize \centering Patches grouped\\by track ID}}
  \put (76.5, 12) {\footnotesize \color{red} 2bc}
  \put (86, 12) {\footnotesize \color{red} fa3}
  \put (98, 12) {\footnotesize \color{red} 1cc}
  \put (112, 23.777) {\parbox[t][8mm][c]{29.533mm}{\footnotesize \centering Filter out patches\\with sun reflections}}
  \put (148.5, 23.477) {\parbox[t][8mm][c]{29.533mm}{\footnotesize \centering Patches indexed\\by plant ID}}
  \put (149.5, 12) {\footnotesize \color{custom-green} $1.1$}
  \put (158.5, 12) {\footnotesize \color{custom-green} $2.3$}
  \put (170.7, 12) {\footnotesize \color{custom-green} $2.9$}
  \put (121.8, 70.5) {\parbox[t][5.2mm][c]{23.024mm}{\footnotesize \centering Track graph}}
  \put (125.7, 61.2) {\tiny 2bc}
  \put (128.4, 61.05) {\tiny 1c9}
  \put (131.4, 60.9) {\tiny 837}
  \put (134.4, 60.8) {\tiny 1a4}
  \put (137.6, 60.6) {\tiny f1f}
  \put (140.2, 60.5) {\tiny bf1}
  \put (124.6, 54.9) {\tiny 6b1}
  \put (127.7, 54.8) {\tiny e18}
  \put (131.4, 54.65) {\tiny fa3}
  \put (134.2, 54.45) {\tiny 20c}
  \put (137.4, 54.4) {\tiny 7bb}
  \put (140.6, 54.25) {\tiny 2ea}
  \put (150.969, 66.518) {\parbox[t][5.2mm][c]{23.024mm}{\footnotesize \centering Plant graph}}
  \put (137.6, 47.982) {\parbox[t][5.2mm][c]{23.024mm}{\footnotesize \centering Match graphs}}
  \put (152, 80.2) {\parbox[t][9.929mm][c]{22.975mm}{\footnotesize \centering Plant file\\(Plant IDs)}}
  \put (121.0, 76.0) {\parbox{21.604mm}{\footnotesize Associate track\\IDs \& plant IDs}}
  \put (150.7, 58.5) {\footnotesize $1.1$}
  \put (156.9, 58.5) {\footnotesize $1.2$}
  \put (163.1, 58.5) {\footnotesize $1.3$}
  \put (169.3, 58.5) {\footnotesize $1.4$}
  \put (150.7, 52) {\footnotesize $2.1$}
  \put (156.9, 52) {\footnotesize $2.2$}
  \put (163.1, 52) {\footnotesize $2.3$}
  \put (169.3, 52) {\footnotesize $2.4$}
  \put (132.5, 39.5) {\footnotesize Track IDs}
  \put (132.9, 33.75) {\footnotesize Plant IDs}
  \put (146.5, 33.75) {\footnotesize \color{custom-green} $1.1$}
  \put (153.1, 33.75) {\footnotesize \color{custom-green} $1.2$}
  \put (162.8, 33.75) {\footnotesize \color{custom-green} $2.9$}
  \put (146.4, 39.5) {\footnotesize \color{red} 2bc}
  \put (152.9, 39.5) {\footnotesize \color{red} 1c9}
  \put (162.6, 39.5) {\footnotesize \color{red} 1cc}
  \end{overpic}
  \caption{Overview of our tool for semi-automatic extraction of PV modules from thermographic videos.}
  \label{fig:overview}
\end{figure*}

This section introduces our tool for semi-automatic extraction of PV modules from thermographic videos. An overview can be found in fig. \ref{fig:overview}. First, the tool splits thermographic videos into individual frames and extracts their GPS coordinates. Aided by the GPS coordinates the user manually specifies which frames belong to which row of the PV plant. PV modules are segmented by Mask R\nobreakdash-CNN, extracted, rectified and stored to disk. A tracking algorithm associates each PV module in subsequent video frames with a unique \emph{track ID}. This way the extracted patches of each PV module can be grouped together. Finally, track IDs are associated with plant IDs. Plant IDs are specified in a standardized \emph{plant file} and describe the electrical wiring and the location of each module in the plant. We chose a semi-automatic approach to achieve a high degree of flexibility and good generalization to different PV plants.

The rest of this section explains the tool in detail.

\subsection{Video Acquisition and Preprocessing}
\label{sec:video_acquisition_and_preprocessing}

Thermographic videos can be captured with any UAV or camera as long as the following requirements are fullfilled:
\begin{itemize}
    \item Each row of the PV plant is scanned individually.
    \item The camera moves monotonically along the row, i.e. there is no significant backward movement.
    \item The current row must be fully visible and always the frontmost (bottommost) one in each frame.
    \item The row must lie approximately horizontal or vertical in each frame.
\end{itemize}
Our tool is robust to changes of the flight velocity, altitude and camera angle. This allows the operator to manually track rows with varying elevation (e.g. hillsides) and choose the optimal camera angle to reduce sun reflections. Additional rows which may become visible in the background due to low camera angles are filtered out.

After acquisition thermal IR videos are split into individual frames and stored as $16$\nobreakdash-bit grayscale TIFFs. The GPS position of each frame is extracted and stored in CSV and KML files. They are needed during the manual grouping of frames that follows in the next step. In case the PV rows are vertical we rotate the video frames by \SI{90}{\degree} to enable equal treatment of both cases in the remaining processing steps.


\subsection{Grouping of Frames into Rows}
\label{sec:grouping_of_frames_into_rows}

\begin{figure}[htbp]
  \centering
  \includegraphics[width=\columnwidth]{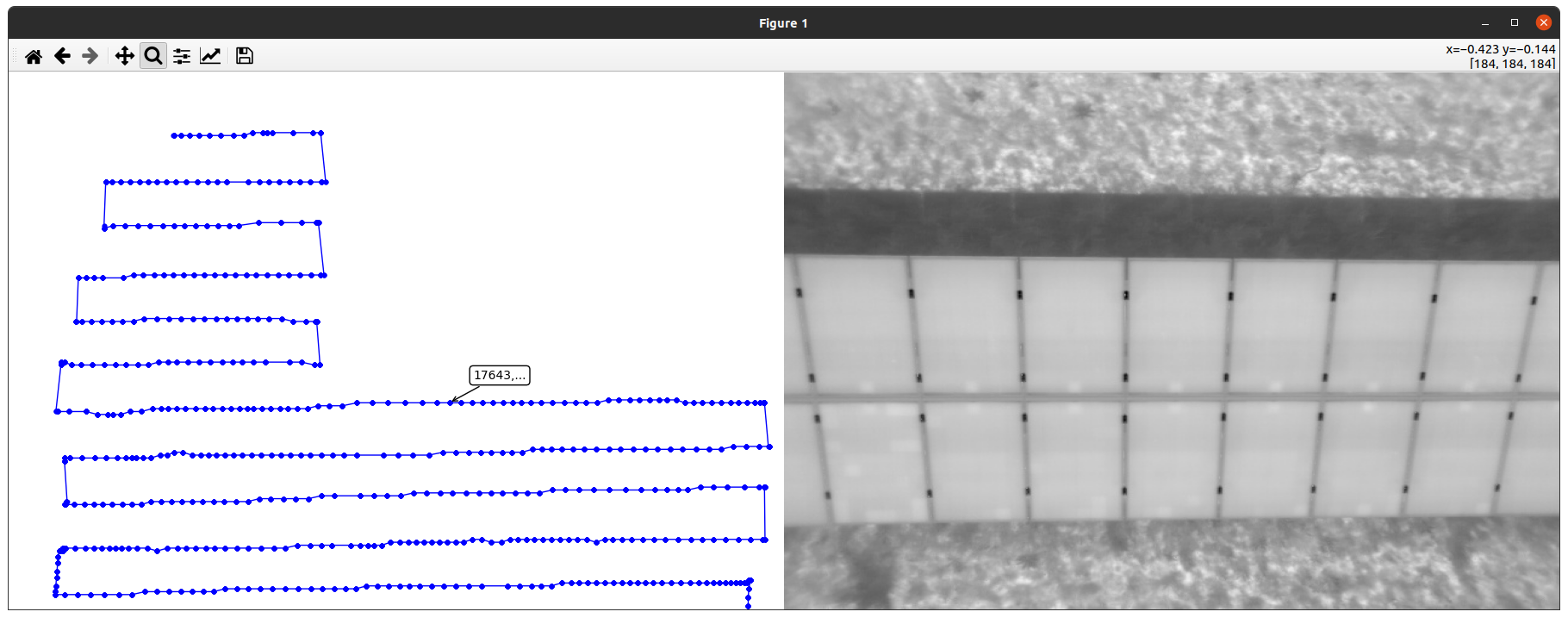}
  \caption{Graphical tool for associating frames with PV plant rows.}
  \label{fig:gps_viewer}
\end{figure}

For maximum flexibility our tool processes each row of the PV plant independently. To this end, the user has to manually specify which video frames belong to which row of the PV plant. Specifically, he has to provide the plant IDs of the bottom left and top right modules and the index of the first and last frame of each row. A graphical tool (see fig. \ref{fig:gps_viewer}) for browsing frames based on their GPS position simplifies this process. The user can skip parts of the video and rows do not need to be scanned in any particular order. It is also possible to scan rows partially, e.g. when a row contains multiple strings of which only a subset needs to be inspected. Further, single frames can be processed which is useful for short rows.

\subsection{PV Module Segmentation}
\label{sec:pv_module_segmentation}

\begin{figure}[htbp]
    \centering
\subfloat{%
     \includegraphics[width=0.32\columnwidth]{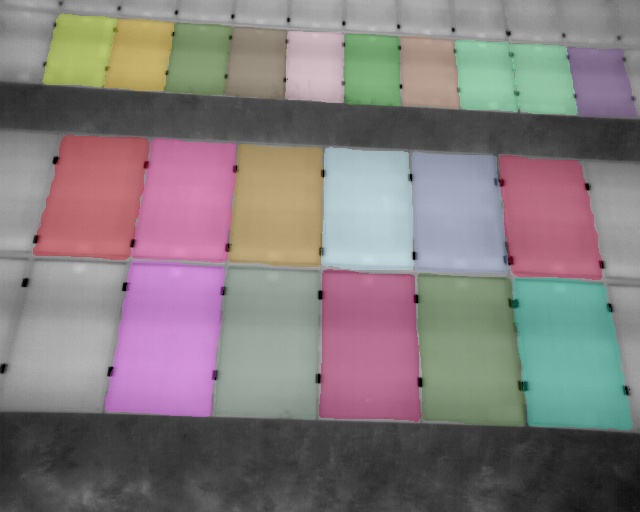}}
\hfill
\subfloat{
        \includegraphics[width=0.32\columnwidth]{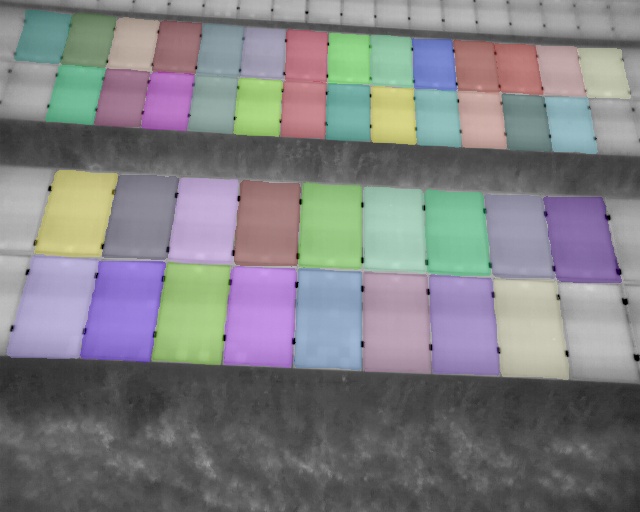}}
\hfill
\subfloat{
        \includegraphics[width=0.32\columnwidth]{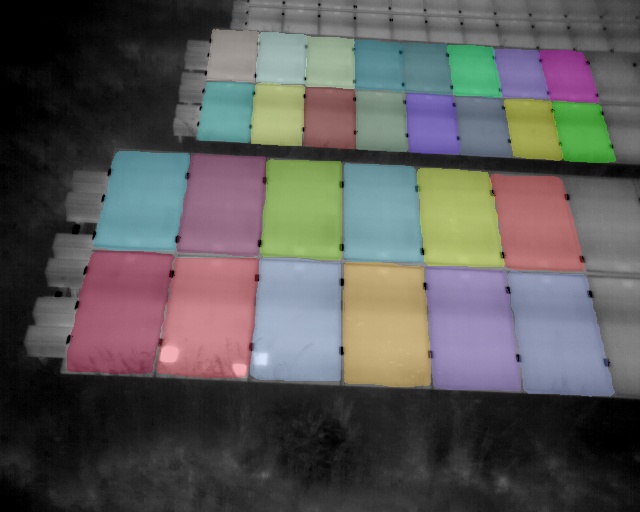}}
  \caption{Example results of the PV module segmentation with Mask R\nobreakdash-CNN.}
  \label{fig:segmentation}
\end{figure}

To locate PV modules in each video frame we use the Mask R\nobreakdash-CNN instance segmentation framework. It outputs an axis-aligned bounding box and a binary segmentation mask for each PV module. We train it to segment only fully visible PV modules. Example outputs are shown in fig. \ref{fig:segmentation}.

\subsubsection{Dataset}
\label{sec:pv_module_segmentation_dataset}
For fine-tuning of Mask R\nobreakdash-CNN we annotate segmentation masks and bounding boxes of $26612$ PV modules in $862$ video frames of PV plants A, B, C and D. For this we developed a custom annotation tool, however any annotation tool for instance segmentation can be used. We select $60$ frames ($15$ of each PV plant) with a total of $2104$ PV modules for validation and the remaining $802$ frames for training. For compatibility with Mask R\nobreakdash-CNN we convert the $16$\nobreakdash-bit grayscale frames to Celsius scale, normalize the values to the interval $[0, 255]$, convert to $8$\nobreakdash-bit, maximize contrast by means of a histogram equalization, convert to RGB and subtract the channel means estimated from the training set. In addition, each frame is padded with zeros to a square of size $640 \times 640$ pixels.

\subsubsection{Training}
Starting from MS COCO-pretrained weights \cite{Lin.2014} we train the segmentation and classification heads of Mask R\nobreakdash-CNN for $59$ epochs using stochastic gradient descent with a batch size of $2$, learning rate $0.001$, momentum $0.9$ and weight decay $0.0001$. Subsequently, all weights are fine-tuned for additional $60$ epochs with $1/10$th of the previous learning rate. During both training stages frames are augmented by random up-down and left-right flips and (in \SI{50}{\percent} of the cases) rotation by a uniform random angle between \SI{-10}{\degree} and \SI{10}{\degree}. We additionally rotate images by \SI{\pm 90}{\degree} in \SI{50}{\percent} of the cases to reduce differences between landscape and portrait orientation of modules.

\subsubsection{Validation Metrics}
We evaluate Mask R\nobreakdash-CNN in terms of F$1$-score and average precision (AP) metric from the MS COCO benchmark \cite{Lin.2014}. To this end, all pairs of predicted and ground truth module bounding boxes in a validation frame are formed and the intersection over union (IoU) of each pair is computed. Pairs with an IoU larger than a specified threshold are true positives ($\textrm{TP}$). False positives ($\textrm{FP}$) are predictions not matched with any ground truth box and false negatives ($\textrm{FN}$) ground truths without predictions. From this, precision $\textrm{TP} / (\textrm{TP} + \textrm{FP})$, recall $\textrm{TP} / (\textrm{TP} + \textrm{FN})$ and F$1$-score $2 \textrm{TP} / (2 \textrm{TP} + \textrm{FP} + \textrm{FN})$ are computed at ten IoU thresholds $\{0.5, 0.55, \ldots, 0.95\}$. AP is the area under the resulting precision recall curve. Finally, F$1$-score and AP are averaged over all validation frames.

\subsubsection{Results}
After fine-tuning Mask R\nobreakdash-CNN achieves an AP of \SI{90.01}{\percent} and an F1-score of \SI{90.51}{\percent}. At IoU threshold $0.5$ the AP and F$1$-score are \SI{99.55}{\percent} and \SI{98.92}{\percent}, respectively. This very good segmentation accuracy allows us to skip any additional filtering and post-processing of the segmentations. Later, in sec. \ref{sec:generalization_of_the_pv_module_segmentation} and \ref{sec:failure_cases} we will analyze how Mask R-CNN generalizes to different PV plants and how segmentation errors affect the PV module extraction.

\subsection{Extraction of Module Patches}

This step extracts segmented PV modules from the thermographic frames and stores the resulting patches to disk. Due to perspective distortion and irregular shape of the segmentation masks direct cropping and storing is not possible. Instead, we fit a minimum-perimeter enclosing quadrilateral to each segmentation mask and obtain a homography which maps the quadrilateral to a rectangle. Width and height of this rectangle correspond to the maximum width and height of the quadrilateral. This yields variable-sized patches which retain most of the information of the source frame without wasting storage space. To ensure each pixel within the quadrilateral is valid we restrict it to lie within the frame. If the IoU of a segmentation mask and the fitted quadrilateral is below $0.9$ the segmentation mask is most likely incorrect and filtered out.


\subsection{PV Module Tracking}

Multiple object tracking is performed to associate segmentation masks of the same PV module in subsequent video frames. This enables grouping of the extracted patches by their associated PV module. To this end, mask centers are projected from frame $t-1$ into frame $t$ using a homography that is estimated by extracting and matching ORB keypoints \cite{Rublee.2011} in both frames. We also tried a Kanade–Lucas–Tomasi tracker but found that it fails due to large motion magnitude whenever the IR camera recalibrates. Each projected mask center is then matched with the nearest segmentation mask center in frame $t$ and its track ID is propagated. If multiple projected mask centers are matched with the same segmentation mask center only the match with the smallest Euclidean distance is considered. The other matches typically correspond to PV modules that left the frame. Whenever a segmentation mask center in frame $t$ is not matched with any of the projected mask centers, a new unique and random track ID is assigned to it. This usually occurs when a new PV module enters the frame.


\subsection{Filtering of the Front Row}
\label{sec:filtering_of_the_front_row}

For low camera angles additional rows of PV modules may be visible in the background of the frame. We develop a filter which discards these background rows and the corresponding patches. It operates independently on each frame and assumes that the currently processed row is the frontmost row (for nadiral videos the bottommost row) in the frame.

\begin{figure}[htbp]
    \centering
\subfloat{%
     \includegraphics[width=0.32\columnwidth]{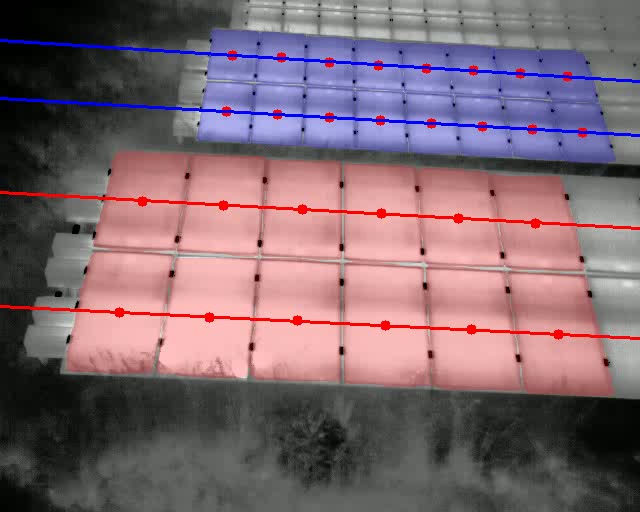}}
\hfill
\subfloat{
        \includegraphics[width=0.32\columnwidth]{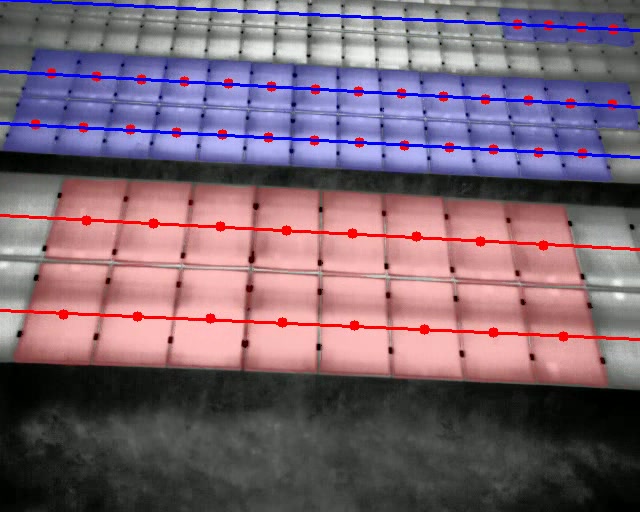}}
\hfill
\subfloat{
        \includegraphics[width=0.32\columnwidth]{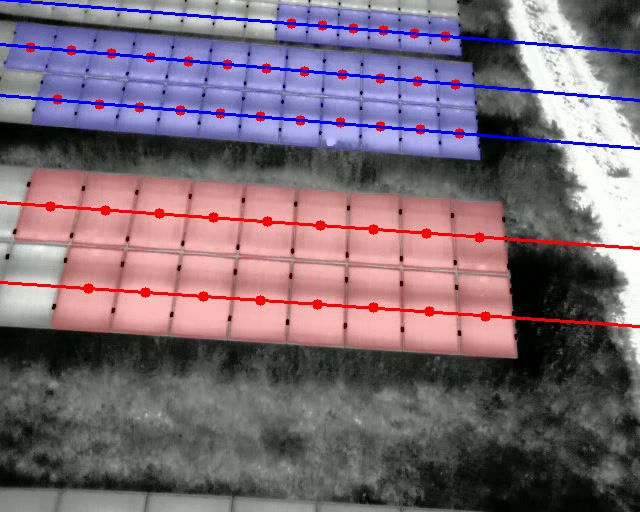}}
  \caption{Result of the front row filtering. Segmentation masks in the front row are colored red, all others blue.}
  \label{fig:row_filtering}
\end{figure}

The filter iteratively fits a line into the set of segmentation mask centers using RANSAC, removes the inlier mask centers and repeats until no more lines can be fit. Each line must deviate at most \SI{\pm 20}{\degree} from the horizontal. During iterative fitting outlier lines can occur which intersect the other lines. We remove them by iteratively removing the line which intersects most other lines until no more intersecting lines are present. Given the number $N$ of vertically stacked PV modules in each row we can retrieve the $N$ lines with largest $y$\nobreakdash-intercept (the image $y$\nobreakdash-axis points downward). The segmentation masks associated with these lines represent the front row and thus are the ones of interest for the further processing steps. Fig. \ref{fig:row_filtering} shows some example outputs of the row filter.

\subsection{Association of Track IDs and Plant IDs}

In this step the random track IDs of PV modules are mapped to plant IDs which encode the electrical wiring of the modules and their location in the plant. The algorithm involves three steps: i) track graph creation, ii) plant graph creation and iii) graph matching.

Both track graph and plant graph encode the spatial relation of all PV modules in a single row of the PV plant. Nodes contain the track IDs and plant IDs, respectively. Edges connect IDs of adjacent modules.

\subsubsection{Track Graph Creation}
The track graph is built iteratively based on all frames associated with the row. For each new frame previously unseen track IDs are added as nodes to the track graph. However, track IDs of spurious tracks (track ID occurring in less than five successive frames) are ignored. Edges are added whenever the overlap, i.e. the number of shared pixels, of two segmentation masks exceeds a threshold. Prior to that all masks are dilated to ensure sufficient overlaps. For PV plants with gaps between module tables adjacent modules are found by additionally searching along a horizontal line passing through the segmentation mask center. In the end, all but the largest connected component of the track graph are removed. The smaller components correspond to background rows resulting from occasional row filtering failures. Additionally, nodes with degree $1$ are removed since they correspond to spurious detections.

\subsubsection{Plant Graph Creation}
Plant graphs are created as one-to-one mappings of the rows in the \emph{plant file} which contain plant IDs and correspond directly to the plant layout.

\subsubsection{Graph Matching}
The final mapping between plant IDs and track IDs of a row is obtained by finding all isomorphisms of the two graphs and selecting the one compatible with a provided seed match between the track ID and plant ID of the bottom left module in the row. The plant ID of this module is provided by the user in an earlier step. Its track ID is found by searching for the bottom left module in the first or last frame of the row using the multi-line fitting approach from above. Whether the first or last frame is used depends on the scan direction (leftward or rightward) which is estimated from the horizontal motion of the tracked modules. As the track graph can contain imperfections an isomorphism can not always be found and instead a subgraph isomorphism is computed. In the seldom case that this also fails the row can not be processed further.


\subsection{Filtering Patches with Sun Reflections}

\begin{figure}[htbp]
  \centering
  \begin{overpic}[width=\columnwidth]{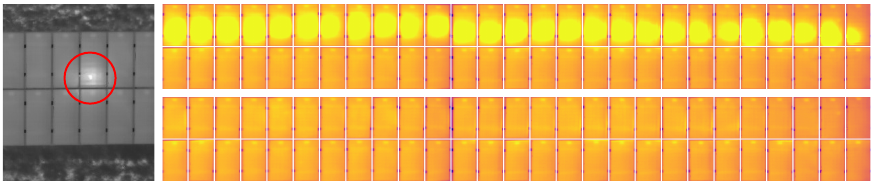}
  \put (1, 4) {\parbox{15mm}{\footnotesize \centering \color{white} Sun\\reflection}}
  \put (20, 12) {\footnotesize No filter}
  \put (20, 1.5) {\footnotesize Filter}
  \end{overpic}
  \caption{Left: Sun reflection in the thermographic video. Right: Extracted patches of a PV plant row with and without the sun reflection filter.}
  \label{fig:sun_filter}
\end{figure}

For some camera angles sun reflections occur which distort the temperature measurement in the thermographic video and the extracted patches (see fig. \ref{fig:sun_filter}). Due to the non-stationary nature of the reflection typically only a subset of the patches of a given PV module is affected. We need to filter them out to prevent issues in the downstream anomaly classification.

The filter finds the maximum temperature $(T_i)_{i=1,\ldots,N}$ and its coordinates $(x_i, y_i)$ in all $N$ subsequent patches of a module. Patches in which $T_i$ and $(x_i, y_i)$ deviate significantly from a reference value most likely contain a sun reflection and are filtered out. More specifically, patch $i$ is filtered out if $|T_i - \bar{T}| > \SI{5}{\kelvin}$ and $\left\lVert (x_i - \bar{x}, y_i - \bar{y}) \right\rVert_2 > \SI{10}{px}$. The reference values $\bar{T}$ and $(\bar{x}, \bar{y})$ are median values computed from a subsequence of the patches which is obtained as follows. First, the discrete difference $p_{i+1} - p_i$ of the Euclidean norm $p_i = \left\lVert (x_i, y_i) \right\rVert_2$ is binarized at a threshold of \SI{10}{px}. All zero-subsequences of $p_i$ which are longer than $0.3 N$ are obtained (the longest is used if none exceeds $0.3 N$). Finally, the zero-subsequence with the smallest variance of the maximum temperature $T_i$ is selected for computation of the reference values.

Fig. \ref{fig:sun_filter} demonstrates the effectiveness of our filter.

\section{Analysis of PV Module Extraction}
\label{sec:analysis_of_pv_module_extraction}

In this section we present the dataset created by our PV module extraction tool and analyze failure cases, processing time and generalization ability.


\subsection{Extracted Dataset}
\label{sec:extracted_dataset}

We run our PV module extraction tool on the seven PV plants in the video dataset and obtain a large-scale dataset with $4.3$ million thermographic patches of $107842$ PV modules (on average $40.0$ patches per module). The tool successfully processed $512$ out of the $561$ PV plant rows (\SI{91.3}{\percent}) and extracted \SI{87.8}{\percent} of all PV modules from the videos. Tab. \ref{tab:extracted_modules_stats} shows details of the extracted dataset and success rates. For plants $E$ and $F$ we use the sun reflection filter which removed $52929$ and $61923$ patches (\SI{6.5}{\percent} and \SI{22.7}{\percent} of the plant total), respectively. The table reports numbers after filtering. Apart from this the same hyper parameters are used for all seven plants indicating a good generalization ability of our extraction tool.

\begin{table*}[htpb]
\centering
\caption{Numbers of PV modules and patches extracted by our tool from the video dataset.}
\label{tab:extracted_modules_stats}
\begin{tabular}{ll
S[table-format=6.0]
S[table-format=5.0]
l
S[table-format=5.0]
l
S[table-format=7.0]
S[table-format=2.1]
}
\toprule
{Plant} & {Sector} & \multicolumn{5}{c}{\# Modules} & \multicolumn{2}{c}{\# Patches}\\\cmidrule(lr){3-7}\cmidrule(lr){8-9}
 & & {Total} & \multicolumn{2}{c}{Extracted} & \multicolumn{2}{c}{Failures} & {Extracted} & {$\varnothing$/Module}\\\midrule
\multirow{5}{1em}{A} & {S0} & 5280 & 5280 & (100.0\,\%) & 0 & (0.0\,\%) & 205488 & 38.9 \\
 & {S1} & 5808 & 5632 & (97.0\,\%) & 176 & (3.0\,\%) & 219653 & 39.0 \\
 & {S2} & 3564 & 3300 & (92.6\,\%) & 264 & (7.4\,\%) & 120100 & 36.4 \\
 & {S3} & 12760 & 11148 & (87.4\,\%) & 1612 & (12.6\,\%) & 430359 & 38.6 \\\cmidrule{2-9}
 & Total & 27412 & 25360 & (92.5\,\%) & 2052 & (7.5\,\%) & 975600 & 38.5 \\\midrule
\multirow{4}{1em}{B} & {S0} & 9297 & 9020 & (97.0\,\%) & 277 & (3.0\,\%) & 232973 & 25.8 \\
 & {S1} & 10990 & 10529 & (95.8\,\%) & 461 & (4.2\,\%) & 370440 & 35.2 \\
 & {S2} & 11478 & 10974 & (95.6\,\%) & 504 & (4.4\,\%) & 364750 & 33.2 \\\cmidrule{2-9}
 & Total & 31765 & 30523 & (96.1\,\%) & 1242 & (3.9\,\%) & 968163 & 31.7 \\\midrule
C &  & 2850 & 2850 & (100.0\,\%) & 0 & (0.0\,\%) & 154476 & 54.2 \\
D &  & 3510 & 2115 & (60.3\,\%) & 1395 & (39.7\,\%) & 128461 & 60.7 \\
E &  & 14688 & 14679 & (99.9\,\%) & 9 & (0.1\,\%) & 766901 & 52.2 \\
F &  & 7280 & 5015 & (68.9\,\%) & 2265 & (31.1\,\%) & 211454 & 42.2 \\
G &  & 35360 & 27300 & (77.2\,\%) & 8060 & (22.8\,\%) & 1107711 & 40.6 \\\midrule
Total &  & 122865 & 107842 & (87.8\,\%) & 15023 & (12.2\,\%) & 4312766 & 40.0 \\\bottomrule
\end{tabular}
\end{table*}




\subsection{Generalization of the PV Module Segmentation}
\label{sec:generalization_of_the_pv_module_segmentation}

In this experiment we analyze how well Mask R\nobreakdash-CNN generalizes to new PV plants. This is practically relevant as fine-tuning on a new plant is time and cost intensive.

To this end, we create training and validation datasets for PV plants A, B, C and D. Validation uses $25$ video frames of each plant, training around $2380$ PV modules per plant. Mask R\nobreakdash-CNN is trained on all combinations of the training sets and its AP (mean of IoU thresholds $\{0.5, 0.55, \ldots, 0.95\}$) is evaluated on each validation set. Training follows sec. \ref{sec:pv_module_segmentation}, however, to speed up the experiment we pretrain and fine-tune for at most $25$ epochs each and always select the model with lowest validation loss.

\begin{figure}[htbp]
  \centering
  \includegraphics[scale=1.0]{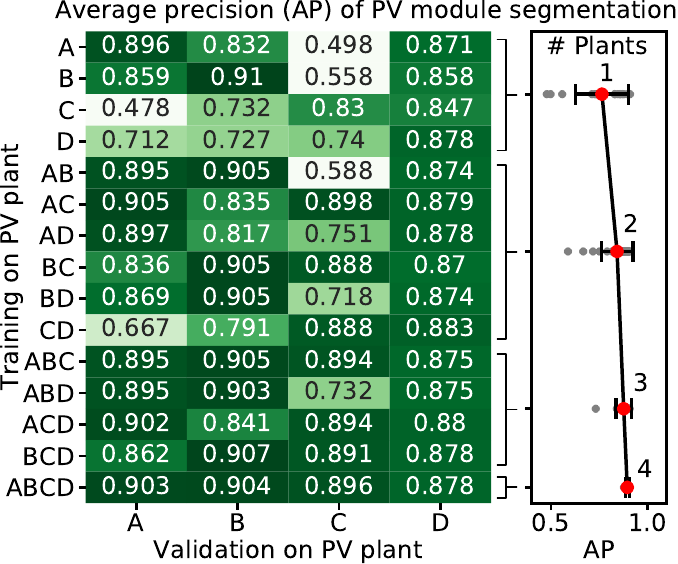}
  \caption{Average precision of the PV module segmentation for all combinations of training sets from PV plants A, B, C and D. The plot on the right shows the mean and standard deviation of the AP when using training data from one, two, three and all four PV plants, respectively.}
  \label{fig:generalization_plot}
\end{figure}

While the results in fig. \ref{fig:generalization_plot} show an increase in validation AP with more training data, they also indicate that plant C differs significantly from plants A, B and D. This is because PV modules are oriented in landscape in plant C and in portrait in plants A, B and D. We validate this by re-running the experiment without randomly rotating frames by \SI{\pm 90}{\degree} during training. This leads to a lower AP of \SI{2.1}{\percent} to \SI{43.7}{\percent} on plant C whenever plant C is not in the training set. Thus, to achieve a high AP Mask R\nobreakdash-CNN must be trained on plant C and at least one of the plants A, B or D. At this point we can not fully explain the low sensitivity of AP for plant D to the training data. We assume distinctive visual features of the PV modules, such as clear boundaries, simplify segmentation.

Fig. \ref{fig:generalization_plot} also reports the mean and standard deviation of all APs when training on one, two, three and four PV plants, respectively. While the standard deviation decreases the mean of the AP increases with more training data. As the AP asymptotically approaches a saturation value the benefit of adding more training data decreases. We found a segmentation model trained on at least three PV plants (of which one is plant C) achieves good results.

\subsection{Failure Cases}
\label{sec:failure_cases}

\begin{figure}[htbp]
     \centering
     \begin{subfigure}[b]{0.24\columnwidth}
         \centering
         \begin{overpic}[width=\textwidth]{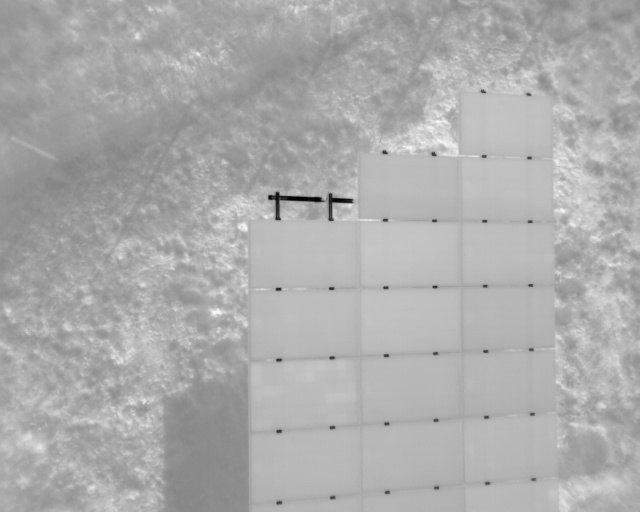}
          \put (5,5) {a}
          \end{overpic}
     \end{subfigure}
     \begin{subfigure}[b]{0.24\columnwidth}
         \centering
         \begin{overpic}[width=\textwidth]{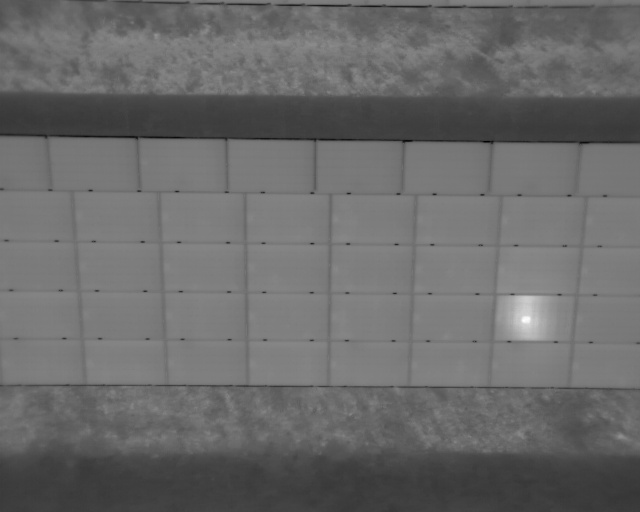}
          \put (5,5) {b}
          \end{overpic}
     \end{subfigure}
     \begin{subfigure}[b]{0.24\columnwidth}
         \centering
         \begin{overpic}[width=\textwidth]{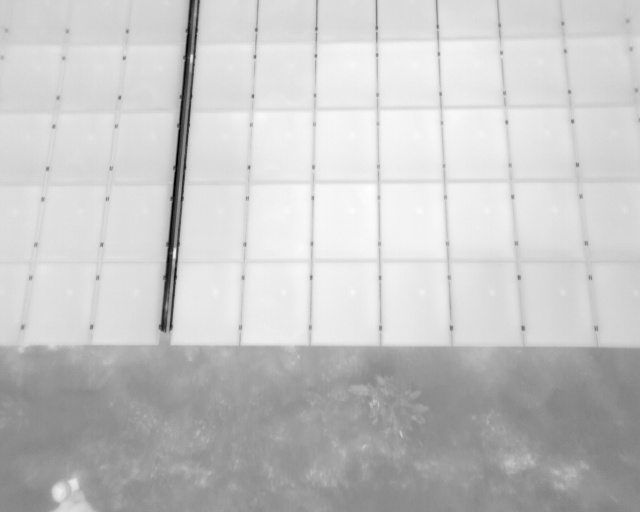}
          \put (5,5) {c}
          \end{overpic}
     \end{subfigure}
     \begin{subfigure}[b]{0.24\columnwidth}
         \centering
         \begin{overpic}[width=\textwidth]{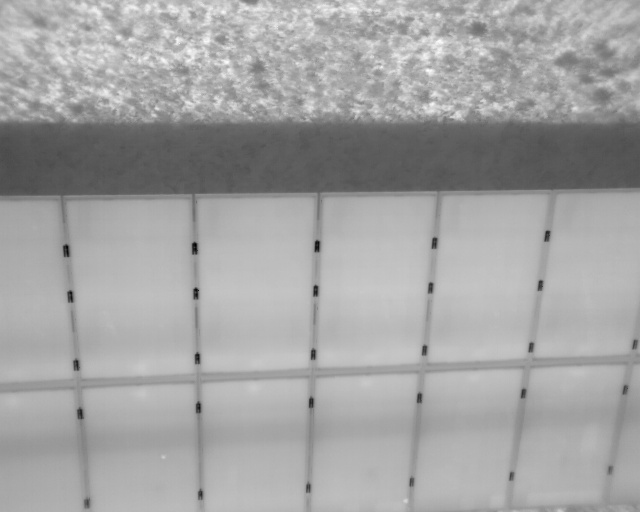}
          \put (5,5) {d}
          \end{overpic}
     \end{subfigure}
     \par\medskip
     \begin{subfigure}[b]{0.24\columnwidth}
         \centering
         \begin{overpic}[width=\textwidth]{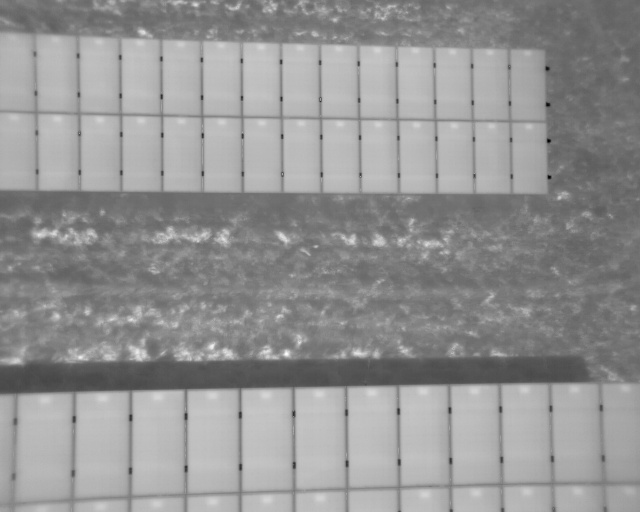}
          \put (5,5) {e}
          \end{overpic}
     \end{subfigure}
     \begin{subfigure}[b]{0.24\columnwidth}
         \centering
         \begin{overpic}[width=\textwidth]{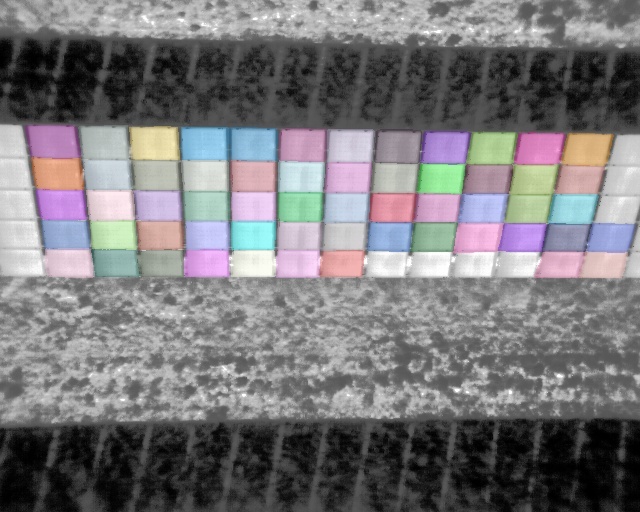}
          \put (5,5) {\color{white} f}
          \end{overpic}
     \end{subfigure}
     \begin{subfigure}[b]{0.24\columnwidth}
         \centering
         \begin{overpic}[width=\textwidth]{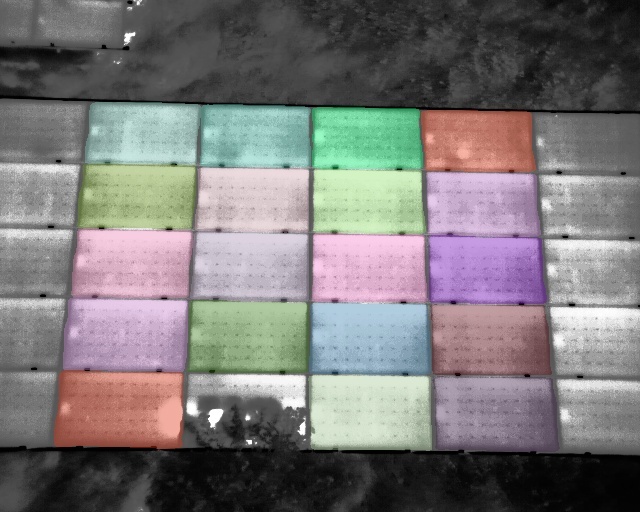}
          \put (5,5) {\color{white} g}
          \end{overpic}
     \end{subfigure}
     \begin{subfigure}[b]{0.24\columnwidth}
         \centering
         \begin{overpic}[width=\textwidth]{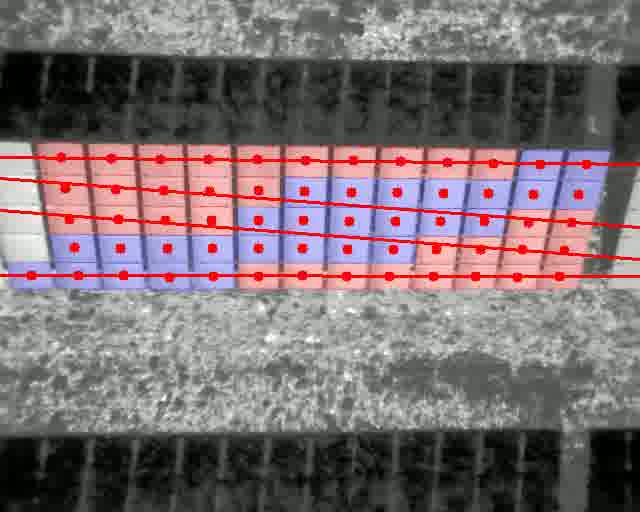}
          \put (5,5) {\color{white} h}
          \end{overpic}
     \end{subfigure}
        \caption{Failure cases of our tool: (a, b) Irregular row layout. (c, d, e) Inadequate UAV trajectory. (f, g) Segmentation error. (h) Row filtering error.}
        \label{fig:failure_analysis}
\end{figure}

Previously, we reported that our tool fails to process $49$ out of $561$ PV plant rows in our video dataset corresponding to \SI{12.2}{\percent} of all PV modules. We identify four common causes: (1) the UAV flight path violates the requirements from sec. \ref{sec:video_acquisition_and_preprocessing}, (2) the PV module segmentation can fail, (3) rows have an irregular layout and (4) the row filtering can fail. Fig. \ref{fig:failure_analysis} shows examples for each failure and tab. \ref{tab:failure_cause_distribution} contains the relative frequencies. We report missed rows instead of missed modules because rows contain varying numbers of modules and an error in a single frame usually leads to loss of the entire row.

\begin{table}[htpb]
\centering
\caption{Numbers of PV plant rows which our extraction tool failed to process.}
\label{tab:failure_cause_distribution}
\begin{tabular}{l
S[table-format=1.0]
S[table-format=1.0]
S[table-format=1.0]
S[table-format=1.0]
S[table-format=1.0]
S[table-format=2.0]
S[table-format=2.0]
S[table-format=2.0]
}
\toprule
Failure Cause & {Plant A} & {B} & {C} & {D} & {E} & {F} & {G} & {All plants}\\\midrule
UAV trajectory & 9 & 2 & 0 & 4 & 2 & 5 & 0 & 22\\
Segmentation error & 0 & 1 & 0 & 0 & 0 & 3 & 10 & 14\\
Irregular row layout & 0 & 4 & 0 & 0 & 0 & 2 & 0 & 6\\
Row filter error & 0 & 0 & 0 & 0 & 0 & 2 & 4 & 6\\
Track graph error & 0 & 1 & 0 & 0 & 0 & 0 & 0 & 1\\\midrule
All failure causes & 9 & 8 & 0 & 4 & 2 & 12 & 14 & 49\\
\bottomrule
\end{tabular}
\end{table}

The majority of rows ($22$ out of $49$) can not be processed due to an inadequate UAV trajectory. This is because some older videos in our dataset were acquired before we established the requirements on the UAV trajectory. Another $14$ rows are missed due to false negatives of the PV module segmentation. They occur mostly in plants $F$ and $G$ on which Mask R-CNN is not fine-tuned and which contain PV modules in landscape orientation. In a few cases segmentation also fails due to sun reflections or occlusion of modules by vegetation. Fine-tuning Mask R\nobreakdash-CNN on more data can mitigate segmentation failures. Irregular row layouts cause failures in six rows. While our tool can handle missing modules some failures still occur because Mask R\nobreakdash-CNN fills gaps in the grid of modules. Further six rows are missed due to failures of the front row filter. They occur only for plants $F$ and $G$ and are related to the lower module segmentation accuracy. A more robust line-fitting method can solve this issue.

For now we tolerate these failures as our extracted dataset is large enough for downstream tasks.

\subsection{Timing Analysis}

\begin{figure}[htbp]
  \centering
  \includegraphics[width=\columnwidth]{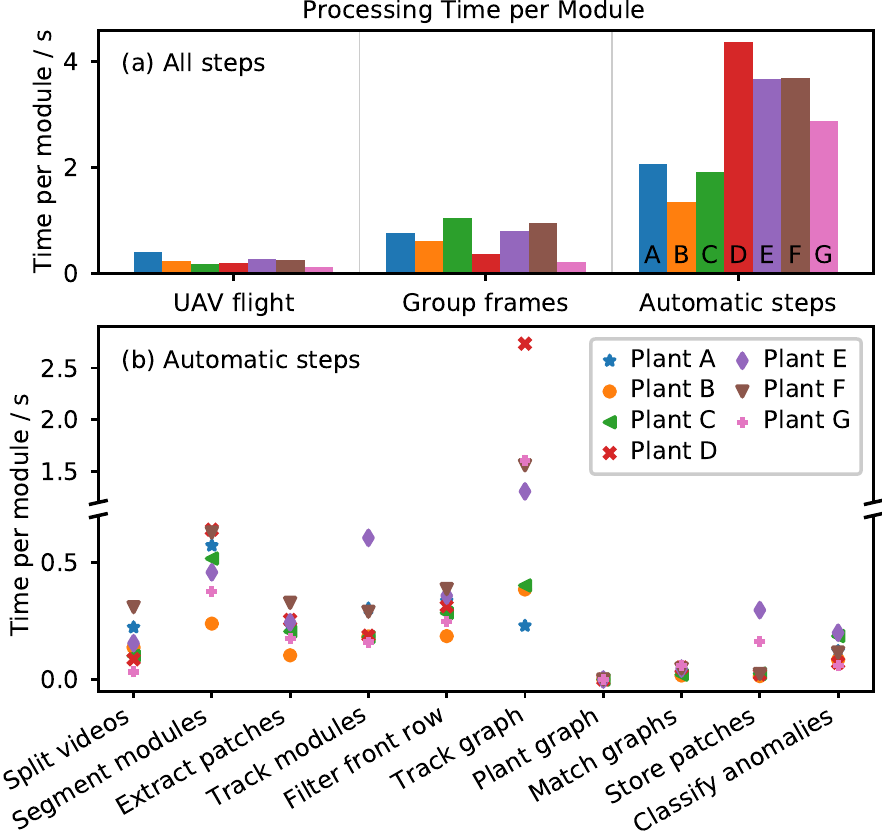}
  \caption{Time needed by our tool to process one PV module. (a) Compares manual and automatic steps. (b) Time distribution of the automatic steps.} 
  \label{fig:timing_analysis}
\end{figure}

Processing time is a critical factor for scaling our tool to larger PV plants. Fig. \ref{fig:timing_analysis} reports timings of both manual and automatic steps of our tool. Automatic steps are timed on a workstation with an Intel Core i9\nobreakdash-9900K, \SI{64}{\giga\byte} of DDR4 RAM, a \SI{4}{\tera\byte} Seagate IronWolf HDD and a GeForce RTX 2080 Ti running Ubuntu 20.04 LTS. Manual steps comprise of UAV flight, frame grouping and plant file creation. The flight duration is estimated from the number of video frames and the frame rate. This underestimates the true duration slightly as battery changes and row changes of the UAV are not considered. For the manual frame grouping we estimate that the user can configure $30$ groups per hour. Due to a lack of accurate measurements fig. \ref{fig:timing_analysis} omits manual plant file creation. It takes $2$ to $8$ hours for a \SI{3}{\mega\watt}$_\textrm{p}$ plant ($10000$ modules) depending on the regularity of its layout.

Timing differences between the plants are due to different video file formats, different plant and row layouts and different UAV flight altitudes and velocities. Track graph creation is faster for plants A, B and C because we can deactivate gap handling. In total, extracting $10000$ modules from a \SI{3}{\mega\watt}$_\textrm{p}$ plant takes $8$ to $21.7$ hours, depending on the plant layout. In here, automatic steps account for $3.8$ to $12.1$ hours which could be significantly reduced by parallelizing the currently sequential processing of PV plant rows. A further speedup is possible by increasing UAV flight velocity and altitude.

\section{Thermal Anomaly Classification}
\label{sec:fault_classification}

In this section we use the extracted thermographic patches for supervised classification of thermal anomalies in PV modules. To this end, we label the patches and train a ResNet\nobreakdash-$50$ classifier to predict whether a patch is nominal or exhibits one of ten common anomalies. As our dataset contains on average $40$ patches per PV module, we choose the majority class across those patches as the final class label for each module.


\subsection{Dataset}
\label{sec:dataset_fault_classification}

\begin{figure}[htbp]
     \captionsetup[subfigure]{aboveskip=2pt, belowskip=0pt, labelformat=empty, justification=centering}
     \newcommand\sizefactor{0.153}
     \centering
     \begin{subfigure}[t]{\sizefactor\columnwidth}
         \centering
         \includegraphics[width=\textwidth]{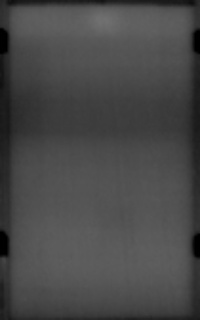}
        \caption{Healthy}
     \end{subfigure}
     \begin{subfigure}[t]{\sizefactor\columnwidth}
         \centering
         \includegraphics[width=\textwidth]{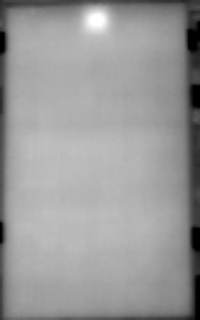}
         \caption{Mh: Module open-circuit}
     \end{subfigure}
     \begin{subfigure}[t]{\sizefactor\columnwidth}
         \centering
         \includegraphics[width=\textwidth]{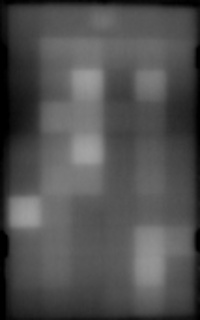}
         \caption{Mp: Module short-circuit}
     \end{subfigure}
     \begin{subfigure}[t]{\sizefactor\columnwidth}
         \centering
         \includegraphics[width=\textwidth]{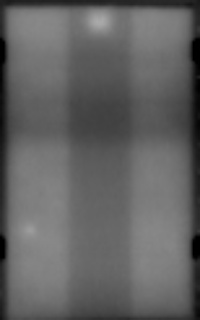}
         \caption{Sh: Substring open-circuit}
     \end{subfigure}
     \begin{subfigure}[t]{\sizefactor\columnwidth}
         \centering
         \includegraphics[width=\textwidth]{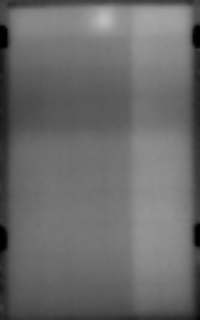}
         \caption{Sh}
     \end{subfigure}
     \begin{subfigure}[t]{\sizefactor\columnwidth}
         \centering
         \includegraphics[width=\textwidth]{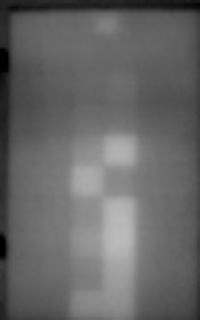}
         \caption{Sp: Substring short-circuit}
     \end{subfigure}
     \par\smallskip
     \begin{subfigure}[t]{\sizefactor\columnwidth}
         \centering
         \includegraphics[width=\textwidth]{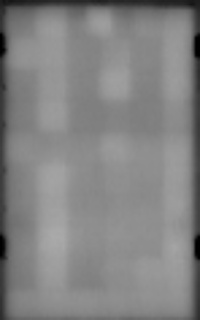}
         \caption{Pid: Module PID}
     \end{subfigure}
     \begin{subfigure}[t]{\sizefactor\columnwidth}
         \centering
         \includegraphics[width=\textwidth]{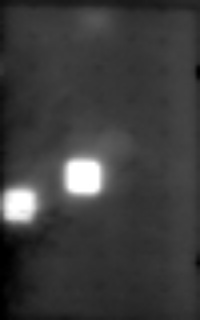}
         \caption{Cm+: Multi. hot cells}
     \end{subfigure}
     \begin{subfigure}[t]{\sizefactor\columnwidth}
         \centering
         \includegraphics[width=\textwidth]{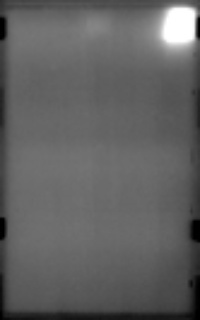}
         \caption{Cs+: Single hot cell}
     \end{subfigure}
     \begin{subfigure}[t]{\sizefactor\columnwidth}
         \centering
         \includegraphics[width=\textwidth]{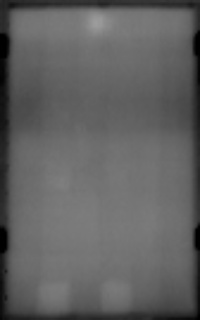}
         \caption{C: Warm cell(s)}
     \end{subfigure}
     \begin{subfigure}[t]{\sizefactor\columnwidth}
         \centering
         \includegraphics[width=\textwidth]{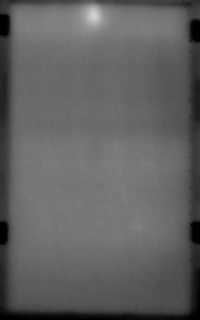}
         \caption{D: Diode overheated}
     \end{subfigure}
     \begin{subfigure}[t]{\sizefactor\columnwidth}
         \centering
         \includegraphics[width=\textwidth]{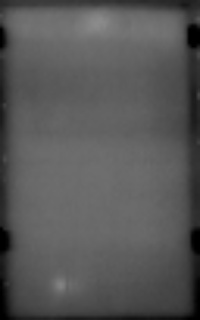}
         \caption{Chs: Hot spot}
     \end{subfigure}
        \caption{Example patches for the ten anomaly classes in our dataset. Severity decreases from left to right and top to bottom. Temperature ranges from \SI{30}{\celsius} (black) to \SI{60}{\celsius} (white). All patches except for class Cm+ are taken from plant A.}
        \label{fig:fault_classification_dataset_examples}
\end{figure}

An expert in our group labels each of the PV modules in our thermographic patch dataset with one out of the ten thermal anomaly classes shown in fig. \ref{fig:fault_classification_dataset_examples}. The class scheme is based on experience and includes relevant module anomalies encountered in previous studies. It is deliberately not optimized for machine learning as the intention is to see how closely the classification of an expert can be reproduced. The structure of our dataset allows to label modules instead of individual patches which speeds up labelling. Note, that we ignore modules of plant D because they are thin-film modules which exhibit different thermal anomalies than the crystalline silicon modules in the other plants. We further exclude all patches with sun reflections from the anomaly dataset and ignore sectors S$1$ and S$2$ of plant B to reduce the labelling workload. To reduce class imbalance (only \SI{6.91}{\percent} of all modules are anomalous) we balance the numbers of healthy and anomalous modules separately for each plant. Finally, we select \SI{70}{\percent} of the PV modules for training, \SI{20}{\percent} for testing and \SI{10}{\percent} for validation. By splitting the data on module-level we ensure that patches of the same module do not occur in multiple splits. The resulting classification dataset (see tab. \ref{tab:fault_classification_dataset}) contains $453511$ patches of $11644$ PV modules half of which are anomalous. There are on average $38.95$ patches per module which act as different augmented views. Note, that the distribution of anomalies differs significantly between the PV plants.

\begin{table*}[htpb]
\centering
\caption{Class distributions of modules and thermographic patches in our anomaly classification dataset.}
\label{tab:fault_classification_dataset}
\begin{tabular}{l
S[table-format=4.0]
S[table-format=4.0]
S[table-format=2.0]
S[table-format=3.0]
S[table-format=4.0]
S[table-format=4.0]
S[table-format=5.0]
S[table-format=6.0]
S[table-format=5.0]
S[table-format=3.0]
S[table-format=5.0]
S[table-format=5.0]
S[table-format=5.0]
S[table-format=6.0]}
\toprule
{Class} & \multicolumn{7}{c}{\# Modules} & \multicolumn{7}{c}{\# Patches}\\\cmidrule(lr){2-8}\cmidrule(lr){9-15}
        & {Plant A} & {B} & {C} & {E} & {F} & {G} & {All plants} & {A} & {B} & {C} & {E} & {F} & {G} & {All plants}\\\midrule
Mh  &    5 &  87 & 4 &   0 &   1 &  494 &  591 &         212 & 2636 & 112 &     0 &    38 & 19968 & 22966\\
Mp  &    2 &   0 & 2 &   5 &   1 &    1 &   11 &          74 &    0 & 151 &   272 &    62 &    26 &   585\\
Sh  &   61 &  31 & 1 &   1 &   1 &    4 &   99 &        2421 &  804 &  43 &    73 &    13 &   145 &  3499\\
Sp  &    9 &   5 & 0 &  33 &   5 &   37 &   89 &         360 &  118 &   0 &  1802 &   217 &  1573 &  4070\\
Pid &  980 & 341 & 0 &   0 &   0 &    0 & 1321 &       40422 & 9143 &   0 &     0 &     0 &     0 & 49565\\
Cm+ &    1 &  10 & 0 &  11 &   6 &    0 &   28 &          26 &  243 &   0 &   477 &   352 &     0 &  1098\\
Cs+ &   12 &  25 & 0 &  11 &  27 &    0 &   75 &         468 &  742 &   0 &   582 &  1348 &     0 &  3140\\
C   &  902 & 184 & 0 & 229 & 570 &    6 & 1891 &       36955 & 4630 &   0 & 11618 & 23539 &   256 & 76998\\
D   &  608 &   1 & 0 &   0 &   3 & 1024 & 1636 &       24891 &   26 &   0 &     0 &   197 & 41210 & 66324\\
Chs &   51 &  17 & 0 &   6 &   1 &    6 &   81 &        1957 &  465 &   0 &   350 &    75 &   205 &  3052\\\midrule
Healthy & 2631 & 701 & 7 & 296 & 615 & 1572 & 5822 &      100725 & 17960 & 302 & 15129 & 25839 & 62259 & 222214\\\midrule
All classes & 5262 & 1402 & 14 & 592 & 1230 & 3144 & 11644 &      208511 & 36767 & 608 & 30303 & 51680 & 125642 & 453511 \\\bottomrule
\end{tabular}
\end{table*}


\subsection{Classifier Training}
\label{sec:clasifier_training}

We initialize ResNet\nobreakdash-$50$ with ImageNet $1.4$M pretrained weights and replace the original fully connected (FC) classification layer with a randomly initialized FC layer containing $11$ neurons. We fix the base model and train only the FC layer for $10$ epochs using Adam optimizer with learning rate $0.001$ and batch size $32$. Afterwards, we fine-tune all layers starting from layer $101$ for another $20$ epochs using RMSprop optimizer with learning rate $1\mathrm{e}{-5}$. During training patches are augmented by random left-right and up-down flips. Preprocessing is similar to the one for segmentation (see sec. \ref{sec:pv_module_segmentation}), however histogram equalization is skipped and patches are resized to $224 \times 224$ pixels without any padding and without maintaining the aspect ratio. During training we do not address class imbalance explicitly.


\subsection{Results}

\subsubsection{Validation Metrics} The ResNet\nobreakdash-$50$ classifier is evaluated on the test set by means of accuracy and per-class F$1$-scores averaged over all classes. Both the unweighted average and the average weighted by class support are reported. We further distinguish patch-level and module-level metrics which are obtained before and after majority voting, respectively. For all metrics we report mean and standard deviation over three training runs.

\subsubsection{Test Performance}

After fine-tuning ResNet\nobreakdash-$50$ achieves \SI{89.40}{\percent} test accuracy on patch-level (see tab. \ref{tab:overall_test_set_metrics}). Majority voting improves it to \SI{90.91}{\percent}. The results are stable over three independent training runs. Training the classifier only on the first patch of each module instead of all patches reduces test accuracy by \SI{5.4}{\percent}. This confirms the benefit of collecting multiple patches per PV module.

\begin{table}[htpb]
\centering
\caption{Test performance of the ResNet\nobreakdash-$50$ classifier on patch- and module-level versus a baseline using only a single patch per PV module.}
\label{tab:overall_test_set_metrics}
\begin{tabular}{l
S[table-format=2.2(3), separate-uncertainty]
S[table-format=2.2(3), separate-uncertainty]
S[table-format=2.2(3), separate-uncertainty]
}
\toprule
 & {Accuracy} & {Unweighted F$1$-score} & {Weighted F$1$-score}\\\midrule
Single patch & 84.00(52) & 58.15(64) & 83.38(55)\\
Patch-level & 89.40(17) & 68.73(106) & 89.18(15)\\
Module-level & 90.91(23) & 70.15(198) & 90.68(24)\\\bottomrule
\end{tabular}
\end{table}

\begin{figure}[htbp]
  \centering
  \includegraphics[scale=1.0]{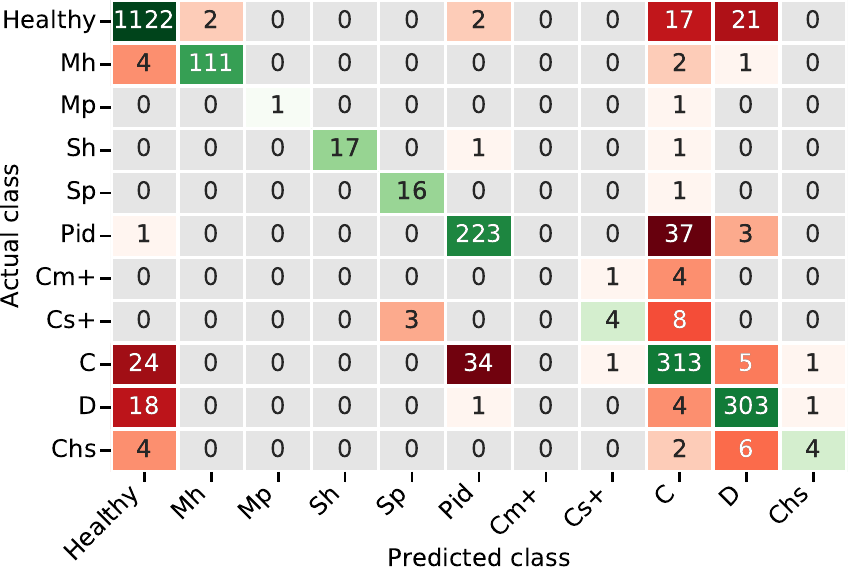}
  \caption{Module-level confusion matrix of the ResNet\nobreakdash-$50$ classifier on the test set. Values are obtained from the first out of three training runs.}
  \label{fig:confusion_matrix}
\end{figure}

As can be seen from the per-class metrics in tab. \ref{tab:per_class_module_level_metrics} and the confusion matrix in fig. \ref{fig:confusion_matrix} the classifier performs well on most anomaly classes, however is less accurate on classes Mp, Cm+, Cs+ and Chs. Reason for this is the under-representation of these classes in our dataset leading to poor generalization from training to test set. Other low-resource classes, such as Sh and Sp, are classified more accurately because the underlying visual patterns are less variable and can be learned accurately from a small number of patches. In some cases, the classifier confuses classes C and D with the healthy modules due to high visual similarity of these classes. Similarly, Pid and C are confused. This is because some Pid modules have comparably little overheated cells and some C modules comparably many of them leading to overlap of the two classes. High visual similarity between some classes also makes labelling difficult and may be a source for considerable amount of noise in the ground truth labels.

\begin{table}[htpb]
\centering
\caption{Per-class module-level metrics of the ResNet-$50$ classifier on the test set. Shown are mean and standard deviation over three training runs.}
\label{tab:per_class_module_level_metrics}
\begin{tabular}{l
S[table-format=3.2(4), separate-uncertainty]
S[table-format=2.2(4), separate-uncertainty]
S[table-format=2.2(4), separate-uncertainty]
S[table-format=5.0]}
\toprule
{Class} & {Precision} & {Recall} & {F$1$-score} & {\# Patches}\\\midrule
Healthy & 95.35(21) & 96.31(19) & 95.83(16) & 1164\\
Mh  & 98.83(42) & 95.76(138) & 97.27(90) & 118\\
Mp  & 66.67(4714) & 33.33(2357) & 44.45(3143) & 2\\
Sh  & 100.00(0) & 87.72(248) & 93.44(142) & 19\\
Sp  & 83.30(76) & 88.24(481) & 85.65(267) &  17\\
Pid & 86.59(175) & 83.71(54) & 85.12(75) & 264\\
Cm+ & 33.33(2357) & 13.33(943) & 19.05(1347) &   5\\
Cs+ & 57.41(693) & 28.89(314) & 38.18(281) &  15\\
C   & 80.39(26) & 83.16(175) & 81.74(97) & 378\\
D   & 90.06(55) & 92.35(43) & 91.19(35) & 327\\
Chs & 57.07(704) & 31.25(510) & 39.75(342) &  16\\\bottomrule
\end{tabular}
\end{table}

\subsubsection{Classifier Visualization}

\begin{figure}[tbp]
     \captionsetup[subfigure]{aboveskip=2pt, belowskip=0pt, labelformat=empty}
     \newcommand\sizefactor{0.153}
     \centering
     \begin{subfigure}[t]{\sizefactor\columnwidth}
         \centering
         \includegraphics[width=\textwidth]{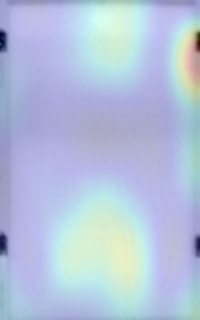}
        \caption{Healthy}
     \end{subfigure}
     \begin{subfigure}[t]{\sizefactor\columnwidth}
         \centering
         \includegraphics[width=\textwidth]{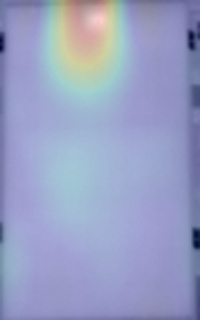}
         \caption{Mh}
     \end{subfigure}
     \begin{subfigure}[t]{\sizefactor\columnwidth}
         \centering
         \includegraphics[width=\textwidth]{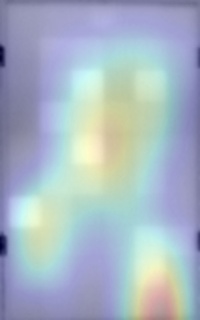}
         \caption{Mp}
     \end{subfigure}
     \begin{subfigure}[t]{\sizefactor\columnwidth}
         \centering
         \includegraphics[width=\textwidth]{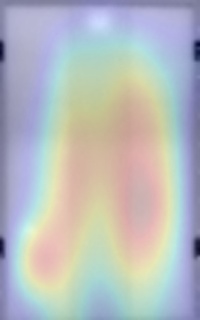}
         \caption{Sh}
     \end{subfigure}
     \begin{subfigure}[t]{\sizefactor\columnwidth}
         \centering
         \includegraphics[width=\textwidth]{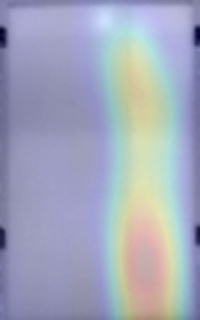}
         \caption{Sh}
     \end{subfigure}
          \begin{subfigure}[t]{\sizefactor\columnwidth}
         \centering
         \includegraphics[width=\textwidth]{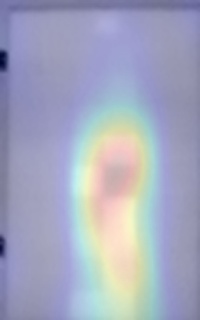}
         \caption{Sp}
     \end{subfigure}
     \par\smallskip
     \begin{subfigure}[t]{\sizefactor\columnwidth}
         \centering
         \includegraphics[width=\textwidth]{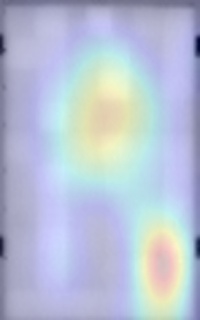}
         \caption{Pid}
     \end{subfigure}
     \begin{subfigure}[t]{\sizefactor\columnwidth}
         \centering
         \includegraphics[width=\textwidth]{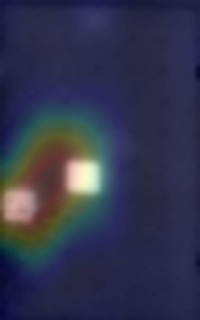}
         \caption{Cm+}
     \end{subfigure}
     \begin{subfigure}[t]{\sizefactor\columnwidth}
         \centering
         \includegraphics[width=\textwidth]{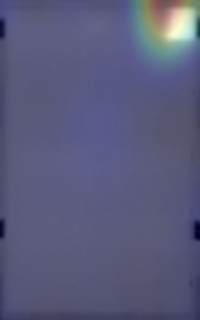}
         \caption{Cs+}
     \end{subfigure}
     \begin{subfigure}[t]{\sizefactor\columnwidth}
         \centering
         \includegraphics[width=\textwidth]{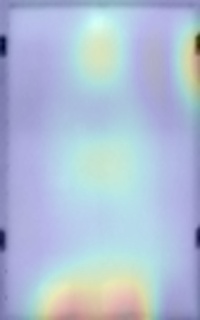}
         \caption{C}
     \end{subfigure}
     \begin{subfigure}[t]{\sizefactor\columnwidth}
         \centering
         \includegraphics[width=\textwidth]{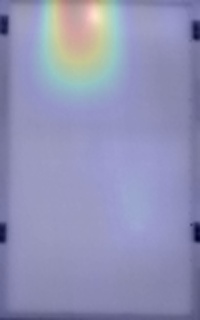}
         \caption{D}
     \end{subfigure}
     \begin{subfigure}[t]{\sizefactor\columnwidth}
         \centering
         \includegraphics[width=\textwidth]{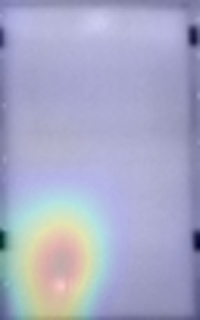}
         \caption{Chs}
     \end{subfigure}
        \caption{Class activation maps of the ResNet\nobreakdash-$50$ classifier obtained with Grad-CAM++ \cite{Chattopadhay.2018}. The patches correspond to fig. \ref{fig:fault_classification_dataset_examples}.}
        \label{fig:class_activation_map}
\end{figure}

To understand if the classifier bases its predictions on meaningful features of the patches we compute class activations maps (CAMs). Fig. \ref{fig:class_activation_map} shows a selection of CAMs. Each CAM visualizes the contribution of a particular image region to the classifier's final prediction. The high correlation between CAMs and temperature anomalies indicates that the classifier draws its confidence mainly from the hot regions in the patch. This is sensible and confirms that the high accuracy of the classifier is based on meaningful image features.

\begin{figure}[tbp]
  \centering
  \includegraphics[scale=0.25]{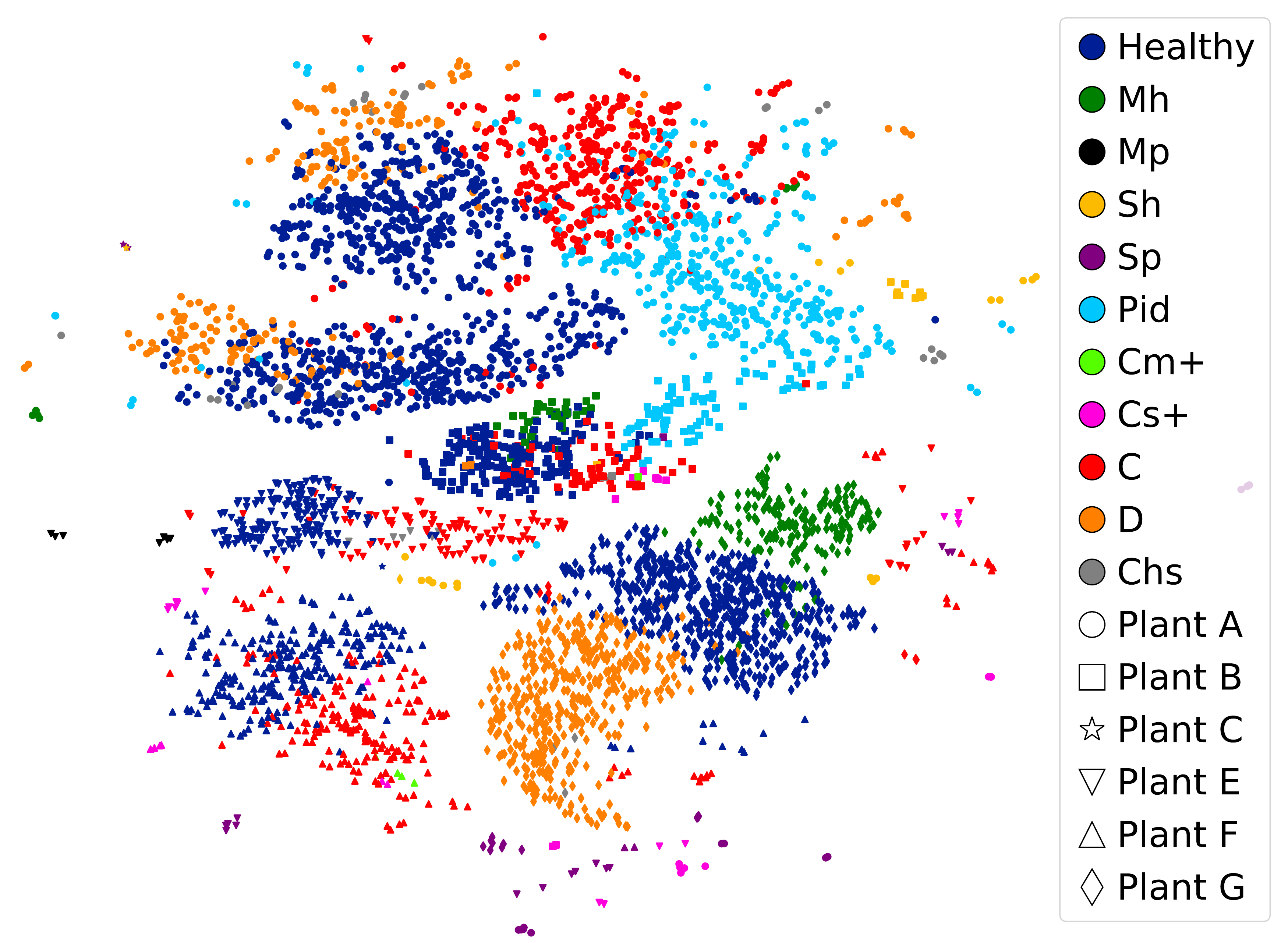}
  \caption{ResNet-$50$ embeddings of the test dataset after dimensionality reduction with UMAP \cite{McInnes.2018}. Embeddings are obtained from the last convolutional layer. Colors represent the ground truth class. For better visualization we show only \SI{5}{\percent} of all data points.}
  \label{fig:embeddings}
\end{figure}

To gain additional insight into the classifier we visualize embeddings of the test set patches in fig. \ref{fig:embeddings}. A few large clusters can be observed which correspond to the six PV plants and most of the anomaly classes. For plant A there are two clusters each because modules in the top row are rotated by \SI{180}{\degree} as compared to those in the bottom row. In addition, several smaller clusters occur which correspond to individual PV modules. Some of them are outliers, others represent classes, such as Cs+ and Sp, which do not form compact clusters due to low sample count and high intra-class variance. The embedding space reflects the classifier's confusion of some classes, e.g. Pid/C and C/D/Healthy, as partial overlap of the respective clusters. Similarly, the low accuracy of some classes, such as Cm+ and Chs, can be explained by the almost complete overlap of the respective clusters with other clusters.

\section{Discussion and Conclusion}
\label{sec:conclusion}

\subsubsection{Summary}
In this work, we developed a computer vision tool for semi-automatic processing of UAV thermographic videos. It handles the large amounts of thermographic images acquired during inspection of PV plants, extracts individual PV modules and classifies ten common module anomalies with an accuracy of more than \SI{90}{\percent} using a ResNet\nobreakdash-$50$ classifier. It further provides the exact location of defective modules in a plant allowing for targeted repairs. Videos are used instead of single images for faster inspection and increased flexibility of UAV operation. Our tool can be used for automated inspection of PV plants superseding an expensive and time-consuming manual inspection. This can reduce cost of PV plant maintenance, ensures safe operation and maximizes yield.

Furthermore, our tool efficiently creates large-scale thermographic datasets by exploiting redundancy in the video. We use this capability to curate a dataset with $4.3$ million thermographic images of $107842$ PV modules from seven PV plants. Modules in the dataset are automatically indexed based on their electrical wiring and location in the plant. This unique index and the large size of the dataset enable research on other downstream machine learning tasks, such as power prediction, which are essential for the safe and profitable operation of future PV plants of ever-growing size.

\subsubsection{State-of-the-art Improvements}
As compared to many of the related works we use deep learning for PV module detection which improves accuracy and generalization. No hyper parameters had to be adjusted to extract modules from the seven different PV plants. By using a deep convolutional classifier for supervised classification of thermal anomalies we followed a recent trend in the field. However, our dataset is significantly larger and we distinguish ten anomaly classes as opposed to at most four classes in the related works. Distinguishing many anomaly classes is not only of value for research datasets but also for plant operators as it facilitates more detailed cataloguing of anomalies in a plant. This is important because some anomalies can worsen over time eventually causing power losses or outages. Despite the larger number of classes test accuracy of our classifier is on par with the related works. However, we also found that classification accuracy is lower for some under-represented classes in our dataset which confirms the need for very large datasets. This also shows that large-scale datasets are required to detect rare anomalies which affect only a handful out of thousands of modules. Smaller datasets as used in many related works do not sufficiently cover such rare anomalies. To allow for even more accurate and fine-grained classification in future we will expand our dataset and explore other deep learning methods which overcome the issue of low accuracy on under-represented classes.

\subsubsection{Future Relevance}
Our work is a first step towards the ultimate goal of automatically characterizing gigawatt-scale PV plants with millions of modules in a day. It shows a way to organize and process the large amounts of data accrued during inspection. However, to achieve full automation and scale up to gigawatt plants multiple UAVs should be used and UAV operation has to be automated. This leads to a predictable scanning order of plant rows which renders most of the manual steps of our tool unnecessary. Scaling up also requires reducing processing time. Given full automation, the worst case throughput of our tool is $19800$ modules per day on a single workstation. To process $3.5$ million modules in a \SI{1}{\giga\watt}$_\textrm{p}$ plant in a day requires a $177$\nobreakdash-fold speedup. This speedup is practically feasible by parallelizing the currently sequential processing of PV plant rows. While this demands for a parallel implementation on a small compute cluster it does not require principle changes to the vision algorithms.

\subsubsection{Future Challenges}
Some challenges remain for future works. For example, the detection of string-level anomalies or faults of non-module components, such as inverters. To this end, multimodal datasets (imagery and electrical) as produced by our tool can be used in combination with machine learning. Future work should also consider additional image sources, such as visual and electroluminescence imagery. For wider applicability anomaly classification could be extended to thin-film, bifacial and half-cell modules, and PV module extraction to plants with non-row layouts, as common in floating PV. Furthermore, methods are needed which predict the PV plant's future health state based on historic data. Finally, the dependency of the anomaly classification on ambient conditions should be explored. We have indications for such a dependency but not yet enough data for a systematic analysis.


\section{Acknowledgements}

The authors would like to thank Janine Denz for valuable discussion about the thermal analysis of PV plants. This work was supported by the Bavarian State Government (project "PV-Tera—Reliable and cost-efficient photovoltaic power generation on the Terawatt scale," no. 44-6521a/20/5). HI ERN gratefully thanks the German Federal Ministry for Economic Affairs and Energy (BMWi) for financial funding of the project COSIMA (FKZ: 032429A) and acknowledge Allianz Risk Consulting GmbH / Allianz Zentrum für Technik (AZT) in Munich, Germany for supporting the project. The authors have declared no conflict of interest.


\FloatBarrier


\end{document}